\documentclass[12pt]{article} 
\pdfoutput=1

\usepackage{amsmath, amsfonts, amssymb}
\usepackage[margin=0.75in]{geometry}
\usepackage{graphicx}
\usepackage[hidelinks]{hyperref}
\usepackage{setspace}
\usepackage{algorithm}
\usepackage{algpseudocode}
\usepackage{tabularx}
\usepackage{lscape} %vertical table
\usepackage{rotating}
\usepackage{hyphenat} 
\usepackage{enumitem}
\usepackage{bm}
\usepackage{placeins} %Added by Ali, to restrict images inside sections/subsections by ``\FloatBarrier'' code
\usepackage{setspace}
\usepackage[capitalise]{cleveref}
\usepackage{fancyhdr}
\usepackage{xcolor}
\usepackage{adjustbox} %Added by Ali, to change the table preferences
\usepackage{tikz} %Added by Ali, just to add checkmark

\usepackage{longtable} %tables spanning multiple pages
\usepackage{caption}

\usepackage{multirow}

\usepackage[flushleft]{threeparttable}

\usepackage{makecell} %table lines in cell

\usepackage{cite}

\usepackage{array, booktabs, makecell} %makecell for breaking table header to multiple lines

\def \bv{\mathbf{u}}
\def \bx{\mathbf{x}}

\definecolor{ForestGreen}{RGB}{34,139,34}

\newcommand{\edit}[1]{\color{black}#1 \color{black}}

\newcommand{\rev}[1]{\color{black}#1 \color{black}}

\title{Interpreting and generalizing deep learning in physics-based problems with functional linear models   }
\author{Amirhossein Arzani$^{1,2}$ \and Lingxiao Yuan$^3$   \and Pania Newell$^1$ \and Bei Wang$^{2,4}$    }

\date{}

\begin{document}

\maketitle

%\newsavebox{\astrutbox}
%\sbox{\astrutbox}{\rule[-5pt]{0pt}{20pt}}
%\newcommand{\astrut}{\usebox{\astrutbox}}

\begin{center}
$^1$Department of Mechanical Engineering, University of Utah, Salt Lake City, UT, USA\\
$^2$Scientific Computing and Imaging Institute, University of Utah, Salt Lake City, UT, USA \\
$^3$Department of Mechanical Engineering, Boston University, Boston, MA, USA.\\
$^4$School of Computing, University of Utah, Salt Lake City, UT, USA\\\
\end{center}

\bigskip

\noindent Correspondence:\\
Amirhossein Arzani,\\
University of Utah,\\
Salt Lake City, UT,  84112\\
Email: amir.arzani@sci.utah.edu
%\newpage

\thispagestyle{empty}

%\doublespacing

%Significance:
%Measuring near-wall blood flow in thin boundary layers is very challenging (numerically and experimentally). Our method for the first time demonstrates that physics-informed machine learning could recover near-wall blood flow from sparse velocity data away from the wall and without even knowing the inlet/outlet boundary conditions. This could provide a new paradigm in modeling near-wall flow when either measurement is difficult or boundary conditions are unknown.

\begin{abstract}

Although deep learning has achieved remarkable success in various scientific machine learning applications, its \edit{opaque} nature poses concerns regarding interpretability and generalization capabilities beyond the training data. Interpretability is crucial and often desired in modeling physical systems. Moreover, acquiring extensive datasets that encompass the entire range of input features is challenging in many physics-based learning tasks, leading to increased errors when encountering out-of-distribution (OOD) data. In this work, motivated by the field of functional data analysis (FDA), we propose generalized functional linear models as an interpretable surrogate for a trained deep learning model. We demonstrate that our model could be trained either based on a trained neural network (post-hoc interpretation) or directly from training data (interpretable operator learning). A library of generalized functional linear models with different kernel functions is considered and sparse regression is used to discover an interpretable surrogate model that could be analytically presented. We present test cases in solid mechanics, fluid mechanics, and transport. Our results demonstrate that our model can achieve comparable accuracy to deep learning and can improve OOD generalization while providing more transparency and interpretability. Our study underscores the significance of \edit{interpretable representation} in scientific machine learning and showcases the potential of functional linear models as a tool for interpreting and generalizing deep learning.


\end{abstract}

\newpage

%%%%%%%%%%%%%%%%%%%%%%%%%%%%%%%%%%%%%%%
%%%%%%%%%%%%%%%%% Introduction %%%%%%%%%%%%%%%
%%%%%%%%%%%%%%%%%%%%%%%%%%%%%%%%%%%%%%%
\section{Introduction} \label{sec:intro}

In recent years, deep learning has emerged as a transformative modeling approach in various science and engineering domains. Deep learning has been successfully used for improving the quality of physical data or improving physics-based models (e.g.,  superresolution~\cite{fukami2023super}, denoising~\cite{Fathietal20}, system/parameter identification~\cite{champion2019data}, and closure modeling~\cite{duraisamy2021perspectives}). Additionally, deep learning is a key tool in machine learning enhanced models where the goal of deep learning is to provide a surrogate for the physics-based model, which is useful in many-query and real-time predictive modeling~\cite{de2020transfer,shukla2023deep}. While deep learning has demonstrated impressive success in most of these studies, its inherent \edit{opaque} nature raises concerns regarding the interpretability of the prediction processes. In physics-based systems, where causal relationships and fundamental first-principle laws play a pivotal role in the results, interpretable models are essential for understanding the phenomena of interest and obtaining trustworthy results. Additionally, it is often desirable for deep learning to generalize and extrapolate beyond the training data once the model is deployed and being used in practice, which is a challenging task in physics-based deep learning~\cite{yuan2022towards}. 

The challenges associated with interpretability and generalization in machine learning and deep learning could be overcome with parsimonious and interpretable models~\cite{kutz2022parsimony}. In physics-based modeling, this has been achieved with various techniques such as symbolic regression~\cite{oh2023genetic}, sparse identification of nonlinear dynamics (SINDy)~\cite{brunton2016discovering}, interpretable reduced-order models (ROM)~\cite{kapteyn2020toward}, and design of certain coordinate transformations in deep neural networks~\cite{champion2019data}. More broadly, the growing field of interpretable and explainable artificial intelligence (XAI) offers a set of tools aimed at making \edit{opaque} deep learning models understandable and transparent to humans~\cite{samek2021explaining,thampi2022interpretable}. XAI approaches could be classified as ``by-design'' and ``post-hoc'' methods. The aforementioned parsimonious models are by-design where one achieves interpretability by building such features in the machine learning model from the initial design phase, which has been a more common approach in  physics-based modeling and scientific machine learning. However, by-design XAI approaches usually lead to a tradeoff between model accuracy and interpretability~\cite{zhong2022explainable}.  On the other hand, post-hoc XAI approaches do not compromise model accuracy and instead, explain the model's results in a post-processing step. Standard off-the-shelf XAI approaches have been recently used in various fields such as healthcare~\cite{ rasheed2022explainable,salahuddin2022transparency}, aerospace~\cite{sutthithatip2021explainable}, turbulence modeling~\cite{saez2022convolutional, kim2023interpretable, cremades2023explaining}, and material science~\cite{zhong2022explainable}. 

Interpretable machine learning models also offer the opportunity to improve generalization. However, generalization to out-of-distribution (OOD) input data is a key challenge in scientific machine learning and particularly for deep learning models~\cite{yuan2022towards}. While standard techniques such as regularization could be used to achieve acceptable in-distribution generalization error (interpolation), OOD generalization (extrapolation) is usually not achieved. Extrapolation poses a serious challenge for \edit{opaque} deep learning models. As an example, machine-learning based turbulence models trained from equilibrium turbulence databases have failed once applied to non-equilibrium turbulence and transitional flows~\cite{fang2022data}. Interestingly, in certain examples, a simple linear regression model has exhibited remarkable performance in extrapolating training data, with an average error rate merely 5\% higher than that of \edit{opaque} models and even surpassed \edit{opaque} models in approximately 40\% of the scientific machine learning prediction tasks evaluated~\cite{muckley2022interpretable}.

Here, we propose a post-hoc deep learning interpretation strategy where we build a surrogate for a given trained neural network in the form of generalized linear integral equations. We hypothesize that the interpretable model also improves OOD generalization while providing an approximation to the neural network's predictions. Our definition of interpretability is based on the work in~\cite{sudjianto2021designing} where interpretability is qualitatively assessed based on characteristics such as additivity, sparsity, and linearity, which are all features of our proposed framework, \edit{as described below.}  Given that many deep learning tasks in scientific computing deal with mapping between functions and functionals, we leverage theories within the field of functional data analysis (FDA)~\cite{horvath2012inference,wang2016functional}. FDA provides a theoretical framework to effectively model and analyze functional data and has been used in different applications~\cite{horvath2012inference, ullah2013applications }. Specifically, we will use functional linear models that enable one to construct analytical mapping involving functions/functionals in the form of interpretable integral equations~\cite{horvath2012inference,wang2016functional }. In scientific machine learning, the learning tasks often involve mapping between high-dimensional data~\cite{arzani2022machine}. In these high-dimensional settings, the simplest interpretable machine learning model, multivariate linear regression, can fail and more advanced interpretable models such as functional regression have been shown to provide better results~\cite{ borggaard1992optimal , griswold2008hypothesis }.  Unlike multivariate methods that discard spatial/temporal distribution of the data, functional methods maintain and leverage the intrinsic structure of the data, capturing the temporal or spatial relationships between data points, and therefore can provide a more accurate mapping between the data and uncover valuable insights and patterns. 

A key challenge in functional regression is the learning of the kernel function that appears in the integral equations. A common approach is expanding the kernel in a certain basis or using a pre-defined fixed kernel~\cite{horvath2012inference,ferratyoxford}. Kernel regression is an established statistical modeling approach~\cite{kohler2014review,ghosh2018kernel} and kernel methods have been used in building nonlinear ROMs~\cite{csala2022comparing, baddoo2022kernel}. In this work, we propose a more flexible framework where the kernel is learned from a library of candidate kernel functions using sparse regression. Once trained on data produced by probing a neural network in a post-hoc fashion, the model will provide an analytical representation in the form of a linear sum of integral equations that not only approximates the neural network's behavior but also provides potential improvement in OOD generalization. The model could be trained based on data probed on the entire training landscape or a subset of the input parameter space to provide a global or local interpretation, respectively.   Our proposed approach could also be viewed in the context of operator learning and neural operators~\cite{kovachki2023neural}. Deep learning of operators has recently gained attention in learning mapping between function spaces and has been utilized in various scientific machine learning problems~\cite{li2020fourier,yin2022simulating,you2022physics,renn2023forecasting  }. Interestingly, certain neural operators also leverage integral equations and generalized versions of functional linear models~\cite{li2020neural}. In scientific computing, the utilization of Green's functions/operators~\cite{duffy2015green,nair2011advanced} has inspired the incorporation of integral equations into the architecture of deep neural operators. These integral equations enable the learning of operators by mapping between function spaces and belong to the category of functional linear models.

In this paper, we present an interpretable machine learning model that builds on several fields such as operator learning, XAI, and FDA.  Our paper provides the following major contributions:
\begin{itemize}
\item We present an early application of functional linear models for post-hoc \edit{interpretable representation} of \edit{opaque} deep learning models in scientific computing.

\item We provide a new library based approach together with sparse regression for discovering the kernels in the functional linear models. This provides more flexibility compared to prior FDA studies with pre-defined kernels.  

\item The majority of post-hoc XAI approaches used in scientific machine learning are local and explain neural network's predictions in a region local to a desired input. Our proposed approach is a global surrogate model that could also be easily adapted to local interpretation tasks.

\item We demonstrate that our proposed functional linear model could be trained either on the data itself or by probing a trained neural network. This allows the model to be utilized either as an interpretable operator learning model or as an \edit{opaque model} interpreter. We document training and OOD testing performance in solid mechanics, fluid mechanics, and transport test cases. 

%\item Our results suggest the notion of a hybrid machine learning strategy where a trained deep learning model is used for in-distribution predictions and an interpretable surrogate is utilized for OOD predictions.

\end{itemize}

The rest of this paper is organized as follows. First, in Sec.~\ref{sec:theo}, we provide a brief theoretical overview of different approaches such as FDA to motivate the use of integral equations as a surrogate for deep learning. Next, we present our proposed functional linear model (Sec.~\ref{sec:flm}) and explain how it is applied for interpretation and OOD generalization in Sec.~\ref{sec:flm2}.  In Sec.~\ref{sec:Res}, we present our results for different scientific machine learning test cases. The results and our framework is discussed in Sec.~\ref{sec:disc}, and we summarize our conclusions in Sec.~\ref{sec:conc}.

%%%%%%%%%%%%%%%%%%%%%%%%%
\section{Methods} \label{sec:met}
\subsection{Theoretical motivation and background }  \label{sec:theo}

Integral equations provide a mathematical framework that encourages the development of interpretable models by explicitly defining the relationships between variables. 
%We motivate their application below. 
Our proposed interpretable surrogate model for understanding a deep learning operator is built upon integral equations. These integral equations yield an interpretable generalized linear model that approximates the predictions of the neural network. We provide a brief review of several topics in applied mathematics and machine learning to motivate the idea of using integral equations to build a surrogate for an available deep learning model. \edit{The theoretical background serves as a motivation for the proposed method and readers may skip to Section~\ref{sec:flm}.}

\subsubsection{Green's functions}

In many physics-based learning tasks, we are interested in solving partial differential equations. Consider  the differential equation $L \mathbf{u}  = \mathbf{f}(\bx)$, where one is interested in solving $ \mathbf{u} $, for different input source terms $\mathbf{f}(\bx)$. Similar to how a linear system of equations $\mathbf{A}\bx=\mathbf{b}$  could be solved as $\bx = \mathbf{A}^{-1}\mathbf{b} $ using an inverse operator $\mathbf{A}^{-1}$, the above differential equation could also be inverted assuming  $L$ is a linear operator 

\begin{equation} 
\mathbf{u}(\bx) = L^{-1} \mathbf{f} = \int  \mathbf{g}(\bx,\bm{\xi}) \mathbf{f}(\bm{\xi})   \,d\bm{\xi} \;,
\label{eqn:green}
\end{equation}
where $  \mathbf{g}(\bx,\bm{\xi})$ is the Green's function corresponding to the linear operator $L$ and the action of $  \mathbf{g}(\bx,\bm{\xi})$  on $\mathbf{f}$ that produces the solution is the Green's operator. Therefore, at least for linear operators one can find an analytical operator representation in the form of an integral equation to map the given input $\mathbf{f}$ to the output $\mathbf{u}$. When dealing with a nonlinear operator, it is
possible to employ a similar concept to find a linear approximation of the operator, at least within a local context. This motivates extending Green's function concept to a generalized linear integral model that can approximate desired physics-based operator learning problems. Given the existing knowledge about Green functions for linear differential equations~\cite{duffy2015green,nair2011advanced}, one can design the integral equations based on the physical problem we are trying to solve. 

\subsubsection{Convolutional neural networks (CNN)} 
Convolutional neural networks (CNN) are arguably one of the most successful deep learning architectures and are widely used in computer vision~\cite{voulodimos2018deep} and mapping 2D image-like field variables in scientific machine learning~\cite{pandey2020perspective,guastoni2021convolutional,zhang2021tonr,fukami2023super}. A key reason behind CNN's success is the fact that each layer is only connected to a local spatial region in the previous layer. This is achieved using convolutional operators that enable CNN to learn hierarchical features. We can write a convolutional integral operation as

\begin{equation} 
\mathbf{u}(x,y) = \int  \mathbf{K}(\zeta, \eta ) \mathbf{f}(x-\zeta,y-\eta)   \,d \zeta\,d\eta=  \int  \mathbf{K}(x - \zeta, y - \eta ) \mathbf{f}( \zeta, \eta)   \,d \zeta\,d\eta \;,
\label{eqn:conv}
\end{equation}
where the output $\mathbf{u}$ is generated by convolving the input $\mathbf{f}$. In CNN, the above operation is done in a discrete manner and the kernel $\mathbf{K}$ represents the learnable parameters of the network. Although convolution in a CNN involves a more complex process of sliding filters across the input and is accompanied by additional operations in different layers, the fundamental idea of a convolutional integral equation that maps inputs to outputs through convolutions inspires the development of integral equation models. Such models can construct interpretable surrogates for CNNs and other deep learning architectures. Interestingly, these convolution layers perform feature learning that once combined with fully connected layers allow CNN to make predictions. Our proposed approach aligns closely with this strategy. Similarly, we leverage a library of integral functions to facilitate feature learning and prediction is made through linear regression. In CNN, the first version of the above equation involving $\mathbf{f}(x-\zeta,y-\eta)$ is used. However, in building our interpretable model, we will use the equivalent version involving $ \mathbf{f}( \zeta, \eta) $ (second form in Eq.~\ref{eqn:conv}).  \edit{Interestingly, a similar analogy between integral equations and neural networks can also be made for fully connected neural networks. The matrix vector multiplications that are building blocks of these networks are known to produce mathematically similar structures to kernel-based integral equations in Eq.~\ref{eqn:green}~\cite{naylor1982linear}.}

\subsubsection{Radial basis function (RBF) networks} 

Radial basis function (RBF) networks are a neural network generalization of kernel regression or classification~\cite{Aggarwal18}.  RBF networks use radial basis functions as their activation function. For a single hidden layer, the output of an RBF network could be written as 

\begin{equation} 
\mathbf{u}(\mathbf{x}) = \sum_{i=1}^{m} w_i \exp(- \frac{ \|  \mathbf{x}- \bm{\mu}_i \|^2 }{ 2 \beta_i^2 } )  \;,
\label{eqn:rbf}
\end{equation}
where $m$ different hidden units with different prototype vector $ \bm{\mu}_i $ and bandwidth $ \beta_i$ are used with $\mathbf{x}$ as an input. The weights of the network $w_i$ are optimized to find the final solution. Each RBF influences a set of points in the vicinity of its feature vector $ \bm{\mu}_i $ with the distance of influence dictated by the bandwidth $ \beta_i$.  RBF networks are universal function approximators. In our library of integral equations for our surrogate model below, we will also leverage RBFs but in the integral form. That is, the feature vector $ \bm{\mu}$ will be replaced with a continuous variable and the integration will be done with respect to this variable. 

\subsubsection{Gaussian process regression (GPR)} 

In Gaussian process regression (GPR), a function is approximated using Gaussian processes, which are specified by a mean function and a covariance function (a kernel)~\cite{williams2006gaussian}. The squared exponential kernel also used in RBF (Eq.~\ref{eqn:rbf}) is a popular choice in GPR. GPR effectively integrates information from nearby points through its kernel function, similar to how we will build our interpretable model below. An intriguing observation is that as the number of neurons in a single hidden layer of a neural network approaches infinity, it evolves into a global function approximator. Similarly, under certain constructs, a neural network with a single hidden layer for a stochastic process converges towards a Gaussian process when the hidden layer contains an infinitely large number of neurons~\cite{williams2006gaussian,neal2012bayesian}.

\subsubsection{Neural operators} 
%paper: Neural Operator: Learning Maps Between Function Spaces With Applications to PDEs

Neural operators are an extension of neural networks that enable  learning of mapping between infinite-dimensional function spaces~\cite{kovachki2023neural,goswami2022physics}. Traditional neural networks also learn a mapping between functions (as used in our test cases below) but they require a fixed discretization of the function, whereas neural operators are discretization-invariant.  In neural operators, typically, each layer is a linear operator (e.g., an integral equation) and nonlinear activation functions are used to increase the expressive power. The input $\mathbf{v}$ to each layer is first passed through an integral linear operator  $ \int  \mathbf{K}(\bx, \bm{\xi}) \mathbf{v}(\bm{\xi})   \,d \bm{\xi}$  using a pre-defined kernel $  \mathbf{K}$ and subsequently a nonlinear activation is applied. Therefore, neural operators also leverage integral equations in their regression tasks but build on neural network architectures for increased expressive power at the price of reduced interpretability. Different designs of the kernel lead to different neural operators. Fourier neural operators (FNO) are a popular and successful example that leverages Fourier transforms and convolutions~\cite{li2020fourier}. Graph neural operators~\cite{li2020neural,huang2023introduction} is another example that uses integral equations similar to the approach we will employ in our model. These operators leverage Monte Carlo sampling techniques to approximate the integral equations.

\subsubsection{Functional data analysis (FDA)} 
FDA is a mathematical framework that focuses on analyzing data in the form of smooth functions, rather than discrete observations~\cite{horvath2012inference,wang2016functional}. We will be presenting our proposed framework within the context of FDA and therefore more information is provided here.  In FDA, the dependent variable, independent variable, or both are functionals. Broadly speaking, we may use FDA to perform mapping and regression when functions are involved either as input or output. Let's consider a mapping between an input functions $\mathbf{f}(\bx)$ and output $\mathbf{u}$, where the output is either a function (scalar/vector field) or a single scalar/vector. In the simplest case mimicking classical regression, for a function output, one might write the output concurrently as $\mathbf{u}(\bx) = \bm{\alpha}(\bx) + \bm{\psi}(\bx)\mathbf{f}(\bx)$, where $\bm\alpha$ and $\bm{\psi}$ are bias and regression coefficient functions, respectively.  However, this simple concurrent formulation does not consider the potential influence of neighboring points on the solution. Integral equations could be used to overcome this issue and provide a more realistic scenario. We can formulate the regression problem using functional linear models~\cite{horvath2012inference,wang2016functional}. Assuming that all data are mean-centered, a fully functional model is applied to the case where the input and output are both functions 
\begin{equation} 
\mathbf{u}(\bx) = \int  \bm{\psi}(\bx, \bm{\xi}) \mathbf{f}(\bm{\xi})   \,d \bm{\xi}   \;, 
\label{eqn:fda1}
\end{equation}
in which the goal is to find $\bm{\psi}$. In a separate problem, when the output is a single scalar/vector value, the problem can be formulated as a scalar/vector response model
\begin{equation} 
\mathbf{u} = \int  \bm{\psi}(\bm{\xi}) \mathbf{f}(\bm{\xi})   \,d \bm{\xi}   \;. 
\label{eqn:fda2}
\end{equation}
Finally, if the output is a function and the input is a single scalar/vector value the problem can be written as a functional response model
\begin{equation} 
\mathbf{u}(\bx)  = \bm{\psi}(\bx) \mathbf{f}   \;. 
\label{eqn:fda3}
\end{equation}
In this paper, we will only study the first two cases (Eq.~\ref{eqn:fda1} and~\ref{eqn:fda2}). It should be noted that Eq.~\ref{eqn:fda3} can be equivalent to a single layer linear feedforward neural network and Eq.~\ref{eqn:fda1}--\ref{eqn:fda2} can be cast as a single layer linear neural operator or a DeepONet. However, the FDA form provides an analytical representation, which assists with interpretation and downstream post-processing. Additionally, as explained below, once combined with sparse regression it allows one to find the appropriate kernel analytically based on data rather than pre-defining it or representing it discretely as in neural networks.

\subsection{Interpretable functional linear models}  \label{sec:flm}
%To improve the expressive power of our model, we can first project the input functions into a higher-dimensional feature space using a set of basis functions (e.g., x^2, x^3,..) and then perform the linear model on this new input variables. 
%badnwidth: a characteristic length scale
%model selection
%FDA ed 2 book: Sec 12.4.4:  Local influence (domain of integration truncated)

The discussion above highlights the importance of integral equations in learning mappings between function spaces. Although the various methods mentioned earlier may have similarities and can be considered equivalent in certain conditions, our primary focus will be on FDA with functional linear models. To enhance the expressive capacity of functional linear models, we will expand their capabilities in three distinct ways:

\begin{itemize}
\item First, we will lift the input functions into a higher-dimensional feature space using a pre-specified lifting map $\mathcal{T}$ (e.g., polynomials) and then define functional linear models for each component of the new feature space separately and use linear superposition to define the final model. Such lifting operations have been successfully used in scientific machine learning models (e.g.,~\cite{qian2020lift}). 

\item We will use generalized functional linear models~\cite{muller2005generalized}. Specifically, we will allow a nonlinear function $g(.)$ to be applied to the functional linear models to create outputs such as $\mathbf{u}(\bx) = g \left( \int  \bm{\psi}(\bx, \bm{\xi}) \mathbf{f}(\bm{\xi})   \,d \bm{\xi} \right)$. 

\item Model selection (choice of the kernel) and tuning its hyperparameters is a difficult task in various forms of kernel regression~\cite{kohler2014review,horova2012kernel}.  Instead of pre-specifying the kernels $ \bm{\psi}$, we will pre-define a library of kernels and associated hyperparameters. Subsequently, we will use sparse regression to select among the library of candidate functions. By specifying the desired level of sparsity, a balance can be achieved between interpretability and accuracy.

\end{itemize}

In the examples explored in this work, we investigate deep learning tasks and corresponding interpretable functional linear models where the input is a 2D function (image) defined on $\Omega$ and the output is either a single scalar value, a 1D function (line), or a 2D function (image). These models can be considered as mappings: $\mathbf{f}(x,y) \to \mathbf{u}$, $\mathbf{f}(x,y) \to \mathbf{u}(x)$, and $\mathbf{f}(x,y) \to \mathbf{u}(x,y)$, respectively. Incorporating the above three modifications to functional linear models and using convolution-like operators for the tasks involving image or line outputs, we write the final models  in the most general form as

\begin{align}
& \mathbf{u}(x,y) =   \sum_{n=1}^{N} \sum_{m=1}^{M} \sum_{\ell=1}^{L}  w_{n,m,\ell} \;   g_n \left(  \int_{\Omega}  \bm{\psi}_m(x-\zeta, y-\eta ) \mathcal{T}_\ell  \mathbf{f}(\zeta,\eta)   \,d \zeta\,d\eta  \right)     \; \; \; \textup{(image to image)}  \;,   \label{eqn:f1} \\
& \mathbf{u}(x) = \sum_{n=1}^{N} \sum_{m=1}^{M} \sum_{\ell=1}^{L}  w_{n,m,\ell} \;   g_n \left(   \int_{\Omega}  \bm{\psi}_m(x-\zeta, \eta ) \mathcal{T}_\ell \mathbf{f}(\zeta,\eta)   \,d \zeta\,d\eta   \right)     \; \; \; \textup{(image to line)}    \;,   \label{eqn:f2} \\
& \mathbf{u} = \sum_{n=1}^{N} \sum_{m=1}^{M} \sum_{\ell=1}^{L}  w_{n,m,\ell} \;   g_n \left( \int_{\Omega}  \bm{\psi}_m(\zeta, \eta ) \mathcal{T}_\ell \mathbf{f}(\zeta,\eta)   \,d \zeta\,d \eta  \right)  \; \; \; \textup{(image to scalar)}   \label{eqn:f3}    \;, 
\end{align}
where a linear combination of $L$ different lifting operations $ \mathcal{T}$ on the inputs, $M$ different kernels $\bm{\psi}$, and $N$ different nonlinear functions $g$ are used in writing the final solution. This could be considered as a generalized version of an additive functional regression~\cite{ferratyoxford,muller2008functional}. \edit{Generalized additive models have been utilized to improve interpretability in deep learning~\cite{agarwal2021neural}.} Our goal is to formulate a linear regression problem based on the above analytical equations and training data to find the unknown coefficients $w_{n,m,\ell} $. We do not impose any constraint on the kernel $\bm{\psi}$ besides being $L^2$, and therefore inducing Hilbert-Schmidt operators. Below we present a few remarks.

\begin{itemize}

\item  The above models are analytically tractable (interpretable), particularly for small $L$, $M$, and $N$. Sparsity promoting regression will be used in this study to eliminate many of the weights $w_{n,m,\ell} $ in a data-driven fashion and improve the interpretability of the final model. The remaining non-zero weights represent a reduced-order representation of the system, which behaves linearly with respect to its parameters $w_{n,m,\ell} $.   

\item In practice, it is not necessary to consider all possible combinations of lifting, kernels, and nonlinearity in the library employed for sparse regression. The library  could be defined in a flexible fashion as an arbitrary combination of these operators and the final solution will be a linear superposition of the selected terms in the library.  

\item The kernels $\bm{\psi}$ provide an interpretation for each term in the model. $  \bm{\psi}(x-\zeta, y-\eta)$ in Eq.~\ref{eqn:f1} represents the effect of input function $\mathbf{f}$ at point ($\zeta$,$\eta$) on the output function $\mathbf{u}$ at point ($x$,$y$). $  \bm{\psi}(x, y)$ in Eq.~\ref{eqn:f3} represents a weight for the influence of the input function $\mathbf{f}$'s value at point  ($x$,$y$) on the output $\mathbf{u}$ and creates a weighted average.

\item Most kernels used are equipped with a bandwidth that also needs to be estimated and represents a characteristic problem-dependent length scale and smoothing parameter. Therefore, in our library of candidate terms, for each such kernel, we also consider several candidate bandwidths and treat each kernel separately. Therefore, $M$ in the above equations is typically a large value. For instance, if three different analytical expressions are proposed for the kernels $\bm{\psi}$ with 20 different potential bandwidths each, then $M=60$.  
%Kernel smoothing MAtlab book Sec 2.4: different methods for selecting h, no universally accepted method. 

\item To enable approximation of the integrals during training, the above integrals are replaced with discrete sums that approximate the integrals. Therefore, the above models could be compared to a graph neural operator with a single hidden layer~\cite{li2020neural}. However, in our model, various kernels are added linearly in parallel to form the final solution in an analytically simple manner, whereas in neural operators the kernels are added sequentially in different hidden layers, which reduces the interpretability.  Additionally, as discussed below, we provide a library approach for kernel selection.

\item In this work, we only study regression tasks. The proposed approach could be extended to classification tasks with appropriate selection of the  nonlinear function $g$~\cite{muller2005generalized}, similar to activation function selection in deep learning. 

\end{itemize}

To find the coefficients $w_{n,m,\ell} $, a linear regression problem is formulated based on the above integral equation models. Let's assume a set of $Q$ training data pairs ($\mathbf{f}$ and $\mathbf{u}$) is available and sampled over a set of collocation points $x_i$ and $y_j$ ($i=1,\dots,I$,  $j=1,\dots,J$) defined on a 2D grid (a total of $N' = I \times J$ points). The input image  $\mathbf{f}(x_i,y_j)$ is mapped to  $\mathbf{u}(x_i,y_j)$,  $\mathbf{u}(x_i)$, or  $\mathbf{u}$ based on the task. Additionally, let's assume a total of P terms is arbitrarily selected among the $L\times M \times N$ candidate terms for the library of integral equations.  The above integral equations could be numerically evaluated using any numerical integration technique for each of the collocation points.  This will result in a system of linear equations in the form $\mathbf{U} = \mathbf{F}\mathbf{W}$, where $\mathbf{U}$ is a $(QN' ) \times 1 $ column vector of outputs, $\mathbf{F}$ is a  $(QN' ) \times P $ regression matrix formed based on evaluating the integrals, and $\mathbf{W}$  is a $ P \times 1$ column vector that contains the unknown coefficients  for each integral equation. Sparse regression is used to find the solution by solving the following convex optimization problem

 \begin{equation}
\min_{ \mathbf{W} } \lVert  \mathbf{U} - \mathbf{F}\mathbf{W} \rVert_2  +  \lambda \lVert \mathbf{W}  \rVert_1 \;,
  \label{eqn:sindy}
\end{equation}
where $\lambda$ is a sparsity promoting regularization parameter. This optimization problem is solved using a sequential thresholded least-squares algorithm~\cite{brunton2016discovering} to find $\mathbf{W}$. Increasing $\lambda$ will reduce the number of active terms in the final integral equation model (improved interpretability) but can reduce the accuracy. Our proposed framework resembles sparse identification of nonlinear dynamics (SINDy) where a similar optimization problem together with a library of candidate terms is used for interpretable data-driven modeling of dynamical systems~\cite{brunton2016discovering}. $\lambda=0.1$ was used for all cases unless noted otherwise. In the Appendix (Sec.~\ref{sec:app}), we present an alternative strategy for solving this linear regression problem by presenting the normal equations for functional linear models. 

The library of candidate terms for each task and test case (defined in the Results Section) is listed in Table~\ref{tab:cand}. The range and number of bandwidths $\beta$ used for each case are also listed. In the more complex tasks, a large number of candidate bandwidths should be selected. Additionally, some of the candidate integral terms were defined based on a truncated domain of integration (local influence), which is a common practice in related methods~\cite{james2009functional,horova2012kernel}.

\begin{table}[h!]
  \caption{The library of equations used for each problem to build an interpretable operator. A sparse regression formulation is used to select among the library of integral equations and form the final analytical solution $u$ by adding the selected terms.}
  \label{tab:cand}
  \centering 
  \begin{threeparttable}
  \begin{scriptsize}
    \begin{tabular}{ccc}
    %{m{15mm} m{70mm} m{18mm}}
    Problem & Library of integral equations \tnote{*} & Notes\\
     \midrule\midrule
\makecell{Image to scalar \\ mapping (case 1 \& 2) \\ $f(x,y) \mapsto u$  }     &   
$\begin{aligned}
&   \iint  f(\zeta,\eta) \,d\zeta\,d\eta  \; \;,      \iint  \zeta f(\zeta,\eta) \,d\zeta\,d\eta  \; \;,          \iint  \eta f(\zeta,\eta) \,d\zeta\,d\eta  \\
&  \iint  \zeta^2 f(\zeta,\eta) \,d\zeta\,d\eta    \; \;,       \iint  \eta^2 f(\zeta,\eta) \,dx\,dy  \; \;,      \iint  \zeta \eta f(\zeta,\eta) \,d\zeta\,d\eta \\
&   \iint  f^2(\zeta,\eta) \,d\zeta\,d\eta  \; \;,     \iint  e^{-(\zeta^2 + \eta^2)/\beta_j} f(\zeta,\eta) \,d\zeta\,d\eta  \; \;,     \iint  e^{-(\zeta^2 + \eta^2)/\beta_j} f^2(\zeta,\eta) \,d\zeta\,d\eta   \\
&  \iint  e^{-\zeta/\beta_j} f(\zeta,\eta) \,d\zeta\,d\eta    \; \;,     \iint  e^{-\eta/\beta_j} f(\zeta,\eta) \,d\zeta\,d\eta   \; \;,     \iint  e^{-f(\zeta,\eta)/\beta_j} \,d\zeta\,d\eta \\
&  (\iint  e^{-(\zeta^2 + \eta^2)/\beta_j} f(\zeta,\eta) \,d\zeta\,d\eta)^2
\end{aligned}
$
 
        &   \makecell{Last term only \\used in case 2. \\ $0.1 <\beta_j  < 10$ \\case 1: $M_\beta$ = 10 \\ \edit{P= 58}  \\case 2: $M_\beta$ = 20 \\ \edit{P= 128}  }                        \\%new row
    \cmidrule(l  r ){1-3}
\makecell{Image to image \\ mapping \\ (case 3 \& 4 \& 6) \\ $f(x,y) \mapsto u(x,y)$  }     &   
$\begin{aligned}
&   \frac{\iint  f(\zeta,\eta) \,d\zeta\,d\eta}{ \iint  \,d\zeta\,d\eta}  \; \;,   \iint  e^{-((x-\zeta)^2 + (y-\eta)^2)/\beta_j} f(\zeta,\eta) \,d\zeta\,d\eta   \; \;,     \iint  e^{-\sqrt{(x-\zeta)^2 + (y-\eta)^2}/\beta_j} f(\zeta,\eta) \,d\zeta\,d\eta    \\
&  \iint  f(\zeta,\eta) \,d\zeta\,d\eta  \mathbf{I}_{2D/\beta_j > 1}(\zeta,\eta)  \; \;,     \iint  f^2(\zeta,\eta) \,d\zeta\,d\eta  \mathbf{I}_{2D/\beta_j > 1}(\zeta,\eta)    \; \;,   e^{\iint  f(\zeta,\eta) \,d\zeta\,d\eta \mathbf{I}_{2D/\beta_j > 1}(\zeta,\eta)}  \\
&    \iint  e^{f(\zeta,\eta)} \,d\zeta\,d\eta  \mathbf{I}_{2D/\beta_j > 1}(\zeta,\eta) \; \;,   \iint  e^{(x-\zeta)/\beta_j} f(\zeta,\eta) \,d\zeta\,d\eta  \; \;,  \iint  e^{(y-\eta)/\beta_j} f(\zeta,\eta) \,d\zeta\,d\eta     \\
&  \tanh ( \iint  e^{(x-\zeta)/\beta_j} f(\zeta,\eta) \,d\zeta\,d\eta )  \; \;,  \tanh (  \iint  e^{(y-\eta)/\beta_j} f(\zeta,\eta) \,d\zeta\,d\eta ) \; \;,    \iint  e^{(x-\zeta)/\beta_j} \tanh f(\zeta,\eta) \,d\zeta\,d\eta   \\
&    \iint  e^{(y-\eta)/\beta_j} \tanh f(\zeta,\eta) \,d\zeta\,d\eta  \; \;,   ( \iint  e^{(x-\zeta)/\beta_j} f(\zeta,\eta) \,d\zeta\,d\eta )^2  \; \;, (  \iint  e^{(y-\eta)/\beta_j} f(\zeta,\eta) \,d\zeta\,d\eta )^2 \\
& \iint  e^{(x-\zeta)/\beta_j} f^2(\zeta,\eta) \,d\zeta\,d\eta  \; \;,    \iint  e^{(y-\eta)/\beta_j} f^2(\zeta,\eta) \,d\zeta\,d\eta \; \;, ( \iint  e^{-((x-\zeta)^2 + (y-\eta)^2)/\beta_j} f(\zeta,\eta) \,d\zeta\,d\eta )^2 \\
& \tanh ( \iint  e^{-((x-\zeta)^2 + (y-\eta)^2)/\beta_j} f(\zeta,\eta) \,d\zeta\,d\eta )
\end{aligned}
$
 
        &   \makecell{case 3:\\ $0.2 <\beta_j  < 1.5$ \\$M_\beta$= 120 \\ \edit{P= 2162}   \\case 4:\\  $0.2 <\beta_j  < 0.4$   \\ $M_\beta$ = 7 \\ \edit{P=128}  \\case 6:\\  $0.2 <\beta_j  < 1.5$   \\ $M_\beta$= 20 \\ \edit{P= 362} }                        \\%new row    \cmidrule(l r ){1-3}\
            \cmidrule(l  r ){1-3}
\makecell{Image to line \\ mapping (case 5) \\ $f(x,y) \mapsto u(x)$  }  &  
  $\begin{aligned}
&   \frac{\iint  f(\zeta,\eta) \,d\zeta\,d\eta}{ \iint  \,d\zeta\,d\eta}   \; \;,   \iint  e^{-((x-\zeta)^2 + \eta^2)/\beta_j} f(\zeta,\eta) \,d\zeta\,d\eta   \; \;,     \iint  e^{-\sqrt{(x-\zeta)^2 + \eta^2}/\beta_j} f(\zeta,\eta) \,d\zeta\,d\eta    \\
&  \iint  f(\zeta,\eta) \,d\zeta\,d\eta  \mathbf{I}_{2D_{wss}/\beta_j > 1}(\zeta)  \; \;,     \iint  f^2(\zeta,\eta) \,d\zeta\,d\eta  \mathbf{I}_{2D_{wss}/\beta_j > 1}(\zeta)    \; \;,   e^{\iint  f(\zeta,\eta) \,d\zeta\,d\eta \mathbf{I}_{2D_{wss}/\beta_j > 1}(\zeta)}  \\
&    \iint  e^{f(\zeta,\eta)} \,d\zeta\,d\eta  \mathbf{I}_{2D_{wss}/\beta_j > 1}(\zeta) \; \;,   \iint  e^{(x-\zeta)/\beta_j} f(\zeta,\eta) \,d\zeta\,d\eta  \; \;,  \iint  e^{-\eta/\beta_j} f(\zeta,\eta) \,d\zeta\,d\eta    \\
&  \tanh ( \iint  e^{(x-\zeta)/\beta_j} f(\zeta,\eta) \,d\zeta\,d\eta )  \; \;,  \tanh (  \iint  e^{-\eta/\beta_j} f(\zeta,\eta) \,d\zeta\,d\eta ) \; \;,    \iint  e^{(x-\zeta)/\beta_j} \tanh f(\zeta,\eta) \,d\zeta\,d\eta   \\
&    \iint  e^{-\eta/\beta_j} \tanh f(\zeta,\eta) \,d\zeta\,d\eta  \; \;,   ( \iint  e^{(x-\zeta)/\beta_j} f(\zeta,\eta) \,d\zeta\,d\eta )^2  \; \;, (  \iint  e^{-\eta/\beta_j} f(\zeta,\eta) \,d\zeta\,d\eta )^2 \\
& \iint  e^{(x-\zeta)/\beta_j} f^2(\zeta,\eta) \,d\zeta\,d\eta  \; \;,    \iint  e^{-\eta/\beta_j} f^2(\zeta,\eta) \,d\zeta\,d\eta \; \;, ( \iint  e^{-((x-\zeta)^2 + \eta^2)/\beta_j} f(\zeta,\eta) \,d\zeta\,d\eta )^2 \\
& \tanh ( \iint  e^{-((x-\zeta)^2 + \eta^2)/\beta_j} f(\zeta,\eta) \,d\zeta\,d\eta )
\end{aligned}
$  
    
       &  \makecell{  case 5:\\ $0.1 <\beta_j  < 1.9$ \\ $M_\beta$ = 120  \\ \edit{P= 2162}   }                           \\% end of rows
    \midrule\midrule
    \end{tabular}
    \begin{tablenotes}
  \item[*]  $\beta_j$ \; $j=1,\dots, M_\beta$ are \edit{uniformly sampled} bandwidth hyperparameters \edit{used for each kernel, and P is the total number of candidate terms in the library.} $\zeta$ and $\eta$ are dummy variables used for integration. $D= \sqrt{(x-\zeta)^2 + (y-\eta)^2} $ and $D_{wss}= \sqrt{(x-\zeta)^2 + \eta^2} $. $\mathbf{I}_A$ is the indicator function for the set $A$ where $\mathbf{I}_A(x,y) = 1$ if $(x,y) \in A$ and  $\mathbf{I}_A(x,y) = 0$ if $(x,y) \notin A$.  $\mathbf{I}_{all}(x,y) = 1$ for all $(x,y)$ and $\mathbf{I}_{all}(x) = 1$ for all $x$. In image-to-image tasks all terms are multiplied by $\mathbf{I}_{all}(x,y)$ and in the image-to-line task by $\mathbf{I}_{all}(x)$ to ensure an image and line are produced as the output, respectively. A constant bias term of  1, $\mathbf{I}_{all}(x,y)$, and  $\mathbf{I}_{all}(x)$ was used in all image-to-scalar, image-to-image, and image-to-line tasks, respectively.

  \end{tablenotes}
 \end{scriptsize}
\end{threeparttable}
  \end{table}

\subsection{Generalizing deep learning with an interpretable surrogate}   \label{sec:flm2}

\begin{figure}[h!]
\centering
\includegraphics[width=0.85\textwidth]{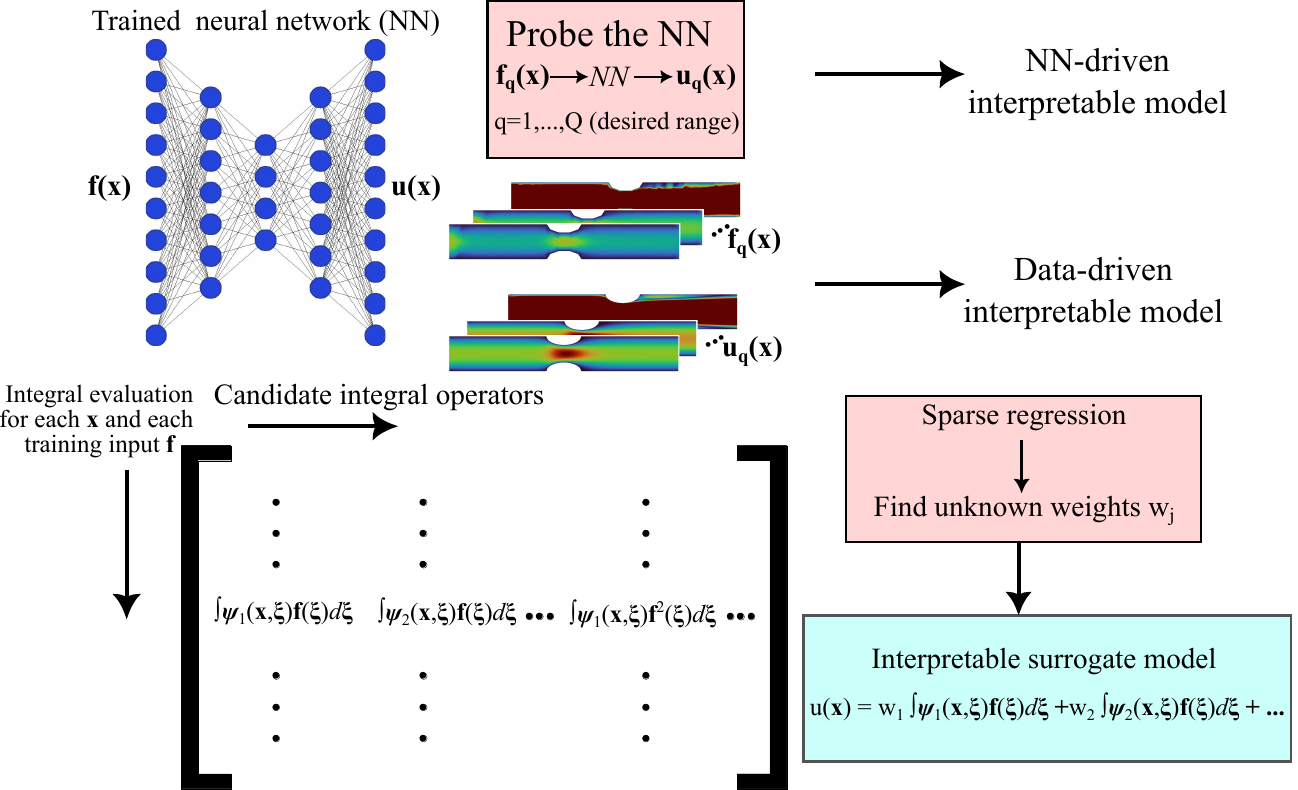}
\caption{An overview of the proposed framework.  Given a trained neural network that maps an input function $\mathbf{f}(\bx)$ to an output function  $\mathbf{u}(\bx)$, the network is probed within a desired range of input data to produce pairs of inputs/outputs. Subsequently, these pairs of data are used to learn an interpretable operator in the form of a linear sum of integral equations (NN-driven interpretable model). Alternatively, the interpretable mode is directly built based on given training data and without a neural network (data-driven interpretable model). The interpretable model is discovered by formulating a sparse regression problem using a library of pre-specified general functional linear models with different kernels.   } 
\label{fig:method}
\end{figure}

Our proposed framework provides an interpretable approach for learning operators and mapping between functions. The entire model is simply a linear combination of integral equations (listed in Table~\ref{tab:cand}). The model is trained by assuming a library of candidate integral equations and solving the convex optimization problem in Eq.~\ref{eqn:sindy}, which allows for the determination of coefficients associated with each integral. Subsequently, given any new input function $\mathbf{f}(x,y)$ one could evaluate the integral equations to find the solution $\mathbf{u}$. The input function's definition is flexible and could be defined either analytically or numerically on an arbitrary grid. A schematic overview is shown in Fig.~\ref{fig:method}.  

\edit{ \textbf{Our definition of interpretability.}  Our definition is based on interpretability features in machine learning such as additivity, sparsity, and linearity as presented in~\cite{sudjianto2021designing}. \rev{These features have also been highlighted in other definitions of interpretable machine learning~\cite{marcinkevivcs2023interpretable,xu2023sparse}.} Our proposed model is additive as different features and terms are added together to find the total analytical equation. Our model is sparse as the number of features in our model (integral equation coefficients) is far fewer than the parameters of a deep neural network. Finally, our model is linear with respect to its unknown parameters. Similar features are used by other studies to assess interpretability. For example, in~\cite{molnar2020quantifying}, a functional decomposition (similar to the additive nature of our model) is used to assess interpretability, where interpretability is assessed with the number of features, interaction strength, and main effect complexity. It should be noted that the first two criteria are inherently imposed by our framework, where we promote sparsity and have zero interaction between features.  
}

In this manuscript, we demonstrate three application areas for our proposed framework:

\begin{enumerate}

\item \textbf{\edit{Interpretable representation of} a trained neural network \edit{(post-hoc analysis).}}  Given a trained neural network for mapping between function spaces, we will probe the network using a desired range of the input function to generate pairs of inputs and outputs. \edit{This is a model-agnostic approach that is independent of the neural network architecture and just depends on the output it generates given each input.} Subsequently, the input and output data will be used to build our interpretable surrogate model, which provides an analytical equation that approximates the behavior of the neural network. The neural network could be probed within the entire range of its training landscape or locally to better understand its behavior in a localized landscape (a specific range of training data). Finally, the network could be probed with out-of-distribution input data to understand the network's behavior outside of its training landscape. It should be noted that the network does not necessarily need to be probed with the exact data that the network used for training. Our definition of interpretation in this work is based on interpretability characteristics proposed in~\cite{sudjianto2021designing} with the goal of demonstrating that neural networks could be approximated with analytical integral equations. Interpreting the physics of the problem using the interpretable model will be future work.   

\item \textbf{Generalizing a trained neural network.}  The surrogate model built based on the data from the probed neural network could also be used to improve out-of-distribution generalization. Namely, the simpler and interpretable model is expected to perform better in extrapolation and generalization. Therefore, one could envision a hybrid model where the neural network is utilized to generate the output when the input data falls within the training landscape. On the contrary, when the input data lies outside of the training landscape, the interpretable surrogate model would be invoked. Of course, this will require one to first determine the boundary of the training landscape, which might not be trivial in some problems~\cite{devries2018learning,yang2021generalized}. 

\item \textbf{An interpretable machine learning model  \edit{(by-design analysis).}}  The interpretable model could be trained directly based on training data to build an interpretable machine learning model in the form of a linear sum of integral equations.

\end{enumerate}

%%%%%%%%%%%%%%%%%%%%%%
\section{Results} \label{sec:Res}

\begin{figure}[h!]
\centering
\includegraphics[scale =0.45]{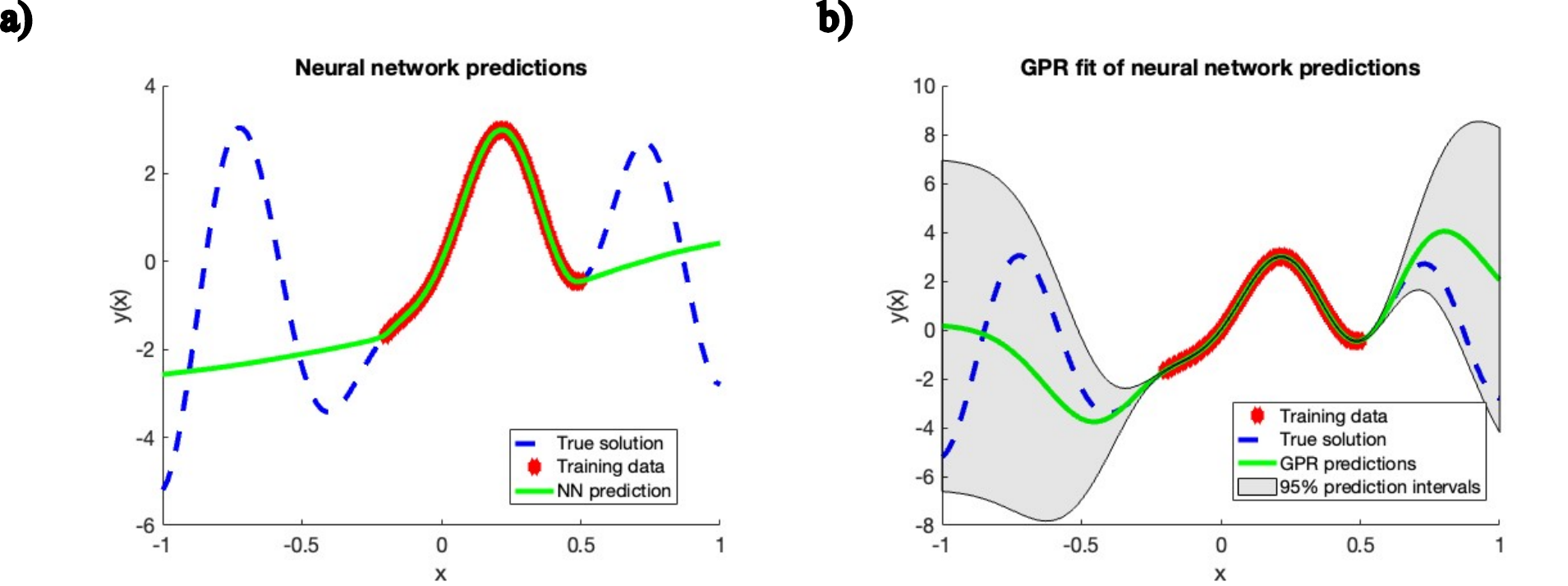}
\caption{A 1D example to motivate the challenge associated with generalization to out-of-distribution input is presented. True data (blue dashed line) and training data (red spheres) are shown. The training data does not cover the entire function.  a) Neural network (NN) prediction. b) Gaussian process regression (GPR) prediction.  The more interpretable GPR model improves prediction for mild extrapolation.   } 
\label{fig:1d}
\end{figure}

First, we will present a simple 1D example to motivate the importance of interpretable machine learning models in the context of generalization. Let's consider the 1D function $u(x) = 4x \sin(11x) + 3\cos(2x)\sin(5x) $. The goal is to learn this function given ($x,u$) training data. We use 120 training points in the range  $-0.2<x<0.5$, which is considered to be the training region. We are interested in observing how the trained machine learning model performs within the range  $-1<x<1$, which will require generalization to out-of-distribution inputs. A fully connected neural network with \edit{one input neuron,} three hidden layers and \edit{35} neurons per layer \edit{(ReLU activations)} and a Gaussian process regression (GPR) model, which is more interpretable than the neural network are used for training. \edit{The neural network was trained with a learning rate of 0.0001 for 6000 epochs using Adam optimization with a weight decay of $10^{-6}$.} The results are shown in Fig.~\ref{fig:1d}. It can be seen that both models perform well within the training region. However, the \edit{opaque} neural network model has worse performance outside of the training region compared to GPR. For mild extrapolation outside the training region, the GPR model has relatively good performance compared to the neural network. \edit{ It should be noted that changing the number of neurons in the neural network model affects the slope of the close-to-linear solution in the extrapolation regime but cannot produce the sinusoidal behavior (results not shown).}

In the following subsections, we will present different examples to test our proposed interpretable model. In each test case, we will quantify the training error and test error. \rev{Validation errors are presented in the Appendix (Sec~\ref{sec:app_valid}).} Throughout the manuscript, by test we imply out-of-distribution test. Errors are quantified for the neural network (NN) model, the interpretable model trained based on the probed trained neural network (Interp NN-driven), and the interpretable model trained based on training data  (Interp data-driven). The mean and maximum errors for each case are listed in Table~\ref{tab:mean} and~\ref{tab:max}, respectively. \rev{Throughout the results, in test cases with mapping to field variables (image-to-image and image-to-line), point-wise absolute error (PAE) aggregates all of the point-wise errors, whereas image-based error calculates the spatially averaged error of each output field variable. The overall mean error is identical between these two approaches since all samples have similar resolution but each approach has different error distributions and maximum errors.}

In cases below, the input data is a 2D scalar field (image) sampled with a 28$\times$28 resolution \edit{and in defining the input field for calculating integrals $ (x, y) \in [0, 1] \times [0, 1]$ was used.} In all cases with the exception of case 1 both input and output fields are normalized. In all examples (except test case 6), the same input training data used in training the neural network was employed for probing the neural network in the NN-driven interpretable model.

\begin{table}[h!]
\centering
\begin{small}
\rev{
\begin{tabular}{ |p{1.8cm}||p{1.8cm}|p{1.8cm}|p{2.5cm}|p{1.7cm}|p{1.8cm}|p{2cm}| }
 \hline
 \multicolumn{7}{|c|}{Mean absolute error (MAE)} \\
 \hline
Test case& NN (train) &  Interp NN-driven (train) & Interp data-driven (train) & NN (test) & Interp NN-driven (test) & Interp data-driven (test)\\
 \hline
case~1 \begin{footnotesize}(MNIST)\end{footnotesize}   & 4.57& 7.54  &   7.29  & 23.98    &12.53 &   8.93  \\
case~1 \; \begin{footnotesize}(EMNIST)\end{footnotesize}  &  9.55 & 9.29&   8.92  & 141.69   &116.88&   90.57 \\
case 2 & 0.0038 & 0.011&   0.011  & 0.17    &0.054&   0.054\\
case 3    & 0.0019  & 0.0057 &   0.0055  & 0.014    &0.0073 &   0.0073\\
case 4    & 0.00058  &  0.0042&   0.0042  & 0.014    &0.027 &   0.024 \\
case 5 &  0.0018  & 0.0018&   0.0013  & 0.18    &0.046&   0.042\\
case 6 &  0.0018     &0.0016 &   0.00017  & --    &--& --\\
 \hline
\end{tabular}
}
 \end{small}
\caption{Mean absolute error (MAE) for the neural network (NN), interpretable model trained on neural network predictions (Interp NN-driven), and interpretable model trained on training data (Interp data-driven) are listed for training and out-of-distribution testing. Test case 6 was based on local interpretation and did not evaluate test data. Additionally, the errors reported for test case 6 are based on the local data used for local evaluation.  }
\label{tab:mean}
\end{table}

\begin{table}[h!]
\centering
\begin{small}
\rev{
\begin{tabular}{ |p{3.3cm}||p{2cm}|p{1.8cm}|p{2.5cm}|p{1.7cm}|p{1.8cm}|p{2.5cm}| }
 \hline
 \multicolumn{7}{|c|}{Maximum absolute error} \\
 \hline
Test case& NN (train) & Interp NN-driven (train) & Interp data-driven (train) & NN (test) & Interp NN-driven (test) & Interp data-driven (test)\\
 \hline
case 1 (MNIST)   & 63.87    &80.04 &   83.28  & 138.68    &56.15 &   30.14  \\
case 1 (EMNIST)  & 72.67    &87.04&   91.70  & 297.44   &294.38&   230.54\\
case 2  & 0.019    &0.11 &   0.11  & 0.37    &0.17 &   0.18\\
case 3 PAE    & 0.10    &0.077 &   0.078  & 0.12    &0.06 &   0.06\\
case 3 image-based    & 0.017    &0.012 &   0.011  & 0.083    &0.022 &   0.021\\
case 4 PAE    & 0.0057    &0.076 &   0.074  &  0.11    &0.18&   0.14\\
case 4 image-based    & 0.0012    &0.0069 &   0.0069  &  0.074    &0.043&   0.038\\
case 5 PAE &  0.015    &0.014 &   0.012  & 0.90    &1.04&   1.19\\
case 5 image-based &  0.0047    &0.0048 &   0.005  & 0.76    &0.18&   0.20\\
case 6 PAE &  0.0077    &0.0064 &   0.0013  & --    &--&   --\\
case 6  image-based &  0.0039    &0.0036 &   0.0003 & --    &--&   --\\
 \hline
\end{tabular}
}
 \end{small}
\caption{ Maximum absolute error for the neural network (NN), interpretable model trained on neural network predictions (Interp NN-driven), and interpretable model trained on training data (Interp data-driven) are listed for training and out-of-distribution testing. In cases where the output is a field, maximum error is either calculated based on point-wise data aggregated across all samples (PAE) \rev{or in an image-based fashion as the spatially averaged error of each output field variable.}  Test case 6 was based on local interpretation and did not evaluate test data. }
\label{tab:max}
\end{table}

%%%%%%%%%%%%%%%%%%%%%%%%%%%%%%%%%%%%%%%%%%%%%%%%%
\subsection{Test case 1: predicting strain energy from a heterogeneous material}

\begin{figure}[h!]
\centering
\includegraphics[width=0.9\textwidth]{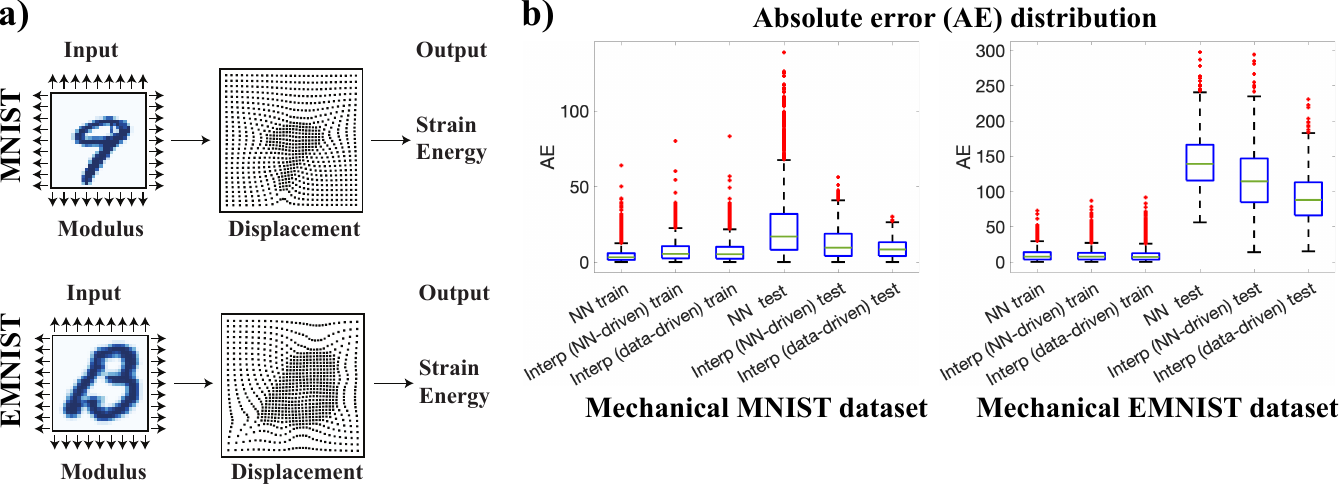}
\caption{Test case 1 results are shown (predicting total strain energy from heterogeneous materials in the Mechanical MNIST and EMNIST datasets). a) An overview of the proposed machine learning task is shown where a single scalar value (strain energy) is predicted from a 2D image (stiffness). b) Boxplots of the absolute error (AE)  distribution are shown. The performance of the neural network (NN), interpretable model trained on neural network predictions (Interp NN-driven), interpretable model trained on training data (Interp data-driven) are shown for the training data and out-of-distribution test data. The AE  boxplot is showing the median (green line), lower/upper quartiles (blue box), the whiskers demonstrate the nonoutlier minimum/maximum of the data, and outliers are shown with red marks. Outliers are defined as values larger than 1.5 times the interquartile range.     } 
\label{fig:t1}
\end{figure}

The Mechanical MNIST--Distribution Shift Dataset~\cite{yuan2022mechanical} consists of finite element simulation data of a heterogeneous material.  As shown in Fig.~\ref{fig:t1}a, the elastic modulus distribution of the heterogeneous material is mapped from the bitmap images of the MNIST and EMNIST datasets~\cite{lecun1998gradient, cohen2017emnist}. The elastic modulus values $E$ of the image bitmaps have non-zero values, and lie within a pre-defined range that depends on the distribution. Pixel bitmaps are transformed into a map of elastic moduli by transforming the pixel value $b$ of the bitmap images through the equation $E= b/255.0*(s-1)+1$. In the Mechanical MNIST--Distribution Shift dataset selected~\cite{lejeune2020mechanical}, the value $s$ is set to 100 for training data and 25 for testing data. In the Distribution Shift EMNIST dataset, \rev{the training data is biasedly sampled with} the value $s$ set to 100 for training data and 10 for testing data. In both cases, equibiaxial extension was applied to the heterogeneous materials through a fixed displacement $d=7.0$ at all boundaries.  In both cases, the training data \edit{size was 2500 and} was randomly split into 80\% training and 20\% validation. A neural network was used to predict the change of strain energy in the material after the extension. The network consists of five fully connected layers \edit{with neurons 1024, 1024, 512, 64, and 1, each followed by a ReLU activation function, except for the final layer. No regularization techniques were applied in test case 1.}  The training data was input as a single batch \edit{(batch size was the size of training data)} and the model was trained at a learning rate 0.001 for 50001 epochs \edit{using Adam optimization.}

The absolute error distribution is shown in boxplots in Fig.~\ref{fig:t1}. Interpretable models improve the test error and the interpretable model trained directly on data has better generalization performance. As also shown in Table~\ref{tab:mean} and~\ref{tab:max}, the two different interpretable model strategies exhibit comparable performance on the training data, and their distinction becomes more apparent during testing. Another notable observation is that, in the case of EMNIST data, the interpretable models exhibit superior \rev{average} performance in training compared to the neural network model and exhibit lower mean errors. However, the improvement is much smaller when considering the improvement in generalization error.

%%%%%%%%%%%%%%%%%%%%%%%%%%%%%%%%%%%%%%%%%%%%%%%%%
\subsection{Test case 2: predicting maximum velocity from a heterogeneous porous medium}

\begin{figure}[h!]
\centering
\includegraphics[width=0.9\textwidth]{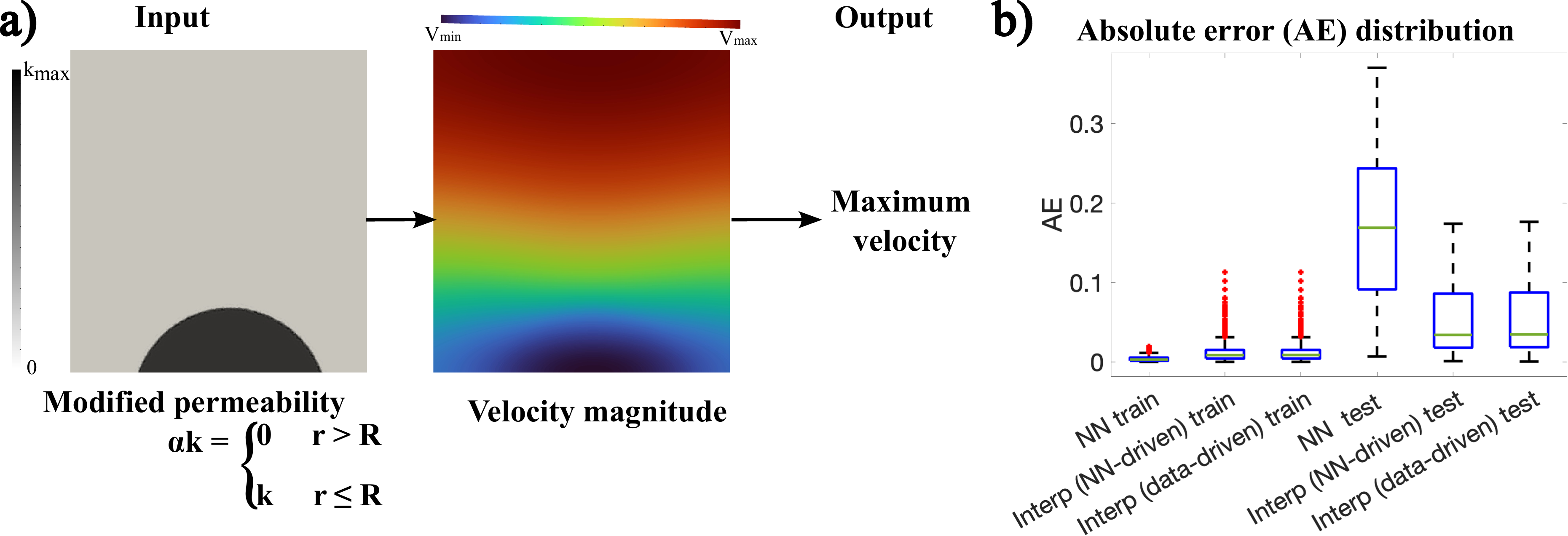}
\caption{Test case 2 results are shown (predicting maximum velocity from permeability fields in porous media flow). a) An overview of the proposed machine learning task is shown where a single scalar value (maximum velocity) is predicted from a 2D image (permeability). b) Boxplots of the absolute error (AE)  distribution are shown. The performance of the neural network (NN), interpretable model trained on neural network predictions (Interp NN-driven), interpretable model trained on training data  (Interp data-driven) are shown for the training data and out-of-distribution test data. Refer to Fig.~\ref{fig:t1} for boxplot details.      } 
\label{fig:t2}
\end{figure}

In this case, we considered porous media flow in a 2D square domain  [0,1] $\times$ [0,1] governed by the steady Darcy-Brinkman equation

\begin{subequations} \label{eq:brink}
\begin{equation}
\alpha \frac{\mu}{k} \bv   = -\nabla p +  \nabla^2 \bv  \;,
\end{equation}
\begin{equation}
\nabla \cdot \bv = 0 \;,
\end{equation}
\end{subequations}
where $\mu=10$  and a heterogeneous permeability of $k(x,y) =0.1\exp(Ax) + 1 $ was used.  Free-slip boundary condition (BC) was imposed at the top and bottom walls (Fig.~\ref{fig:t2}a) and the flow was driven by a pressure gradient (p=1 and p=0 on the left and right sides, respectively). The porous domain was switched on using the $\alpha$ parameter set to  $\alpha=1$ when $\sqrt{(x-0.5)^2 + (y-Y)^2 } \leq  R $ and $\alpha =0$ otherwise as shown in Fig.~\ref{fig:t2}a.  Training data was generated by varying $A$, $Y$, and $R$ within   $0 \leq A \leq 2 $,   $-0.1 \leq Y \leq 0.15 $, and  $0.09 \leq R \leq 0.16 $. The goal of the deep learning model was to predict maximum velocity $\bv$ given $\alpha k(x,y)$ as the input function.  A total of 2250 2D simulations were performed using the open-source finite-element method solver FEniCS~\cite{LoggMardalWells12} using $\sim$70k triangular elements. The data were randomly split into \rev{80\% training and 20\% validation.}  Out-of-distribution test data was also generated by running 100 simulations within  $0 \leq A \leq 2 $,   $0.2 \leq Y \leq 0.3 $, and  $0.1225 \leq R \leq 0.2025 $ (note that $Y$ is completely outside the previous range). A convolutional neural network with three layers of convolution (5$ \times$5 kernel, 6,16,32 channels, and maxpooling \edit{after the second and third layers)} was used followed by three hidden fully connected layers to map the input 2D function into a single scalar value. \edit{ReLU activation functions were used.} 2000 epochs with a learning rate of 5$\times$ 10$^{-4}$ and  a batchsize of 64 were used. \edit{Stochastic gradient descent optimization was used with a $10^{-6}$ weight decay.} In this example, the L1 regularized formulation (Eq.~\ref{eqn:sindy}) did not produce good test results compared to the neural network, and therefore an L2 regularization was used (presented in the Appendix, Sec.~\ref{sec:app}). $\lambda = 10^{-9}$ was the L2 regularization parameter and the preconditioned conjugate gradients method was used for solving the normal equations. 

The absolute error distribution is shown in boxplots in Fig.~\ref{fig:t2}b. In this case, as expected the neural network had a better training error compared to the interpretable models. However, the interpretable models significantly reduce the test error. In this case, the NN-driven and data-driven interpretable models had similar performance in training and testing, which is likely due to the very good neural network training error.

%%%%%%%%%%%%%%%%%%%%%%%%%%%%%%%%%%%%%%%%%%%%%%%%%
\subsection{Test case 3: predicting velocity magnitude field from a heterogeneous porous medium}

\begin{figure}[h!]
\centering
\includegraphics[width=0.9\textwidth]{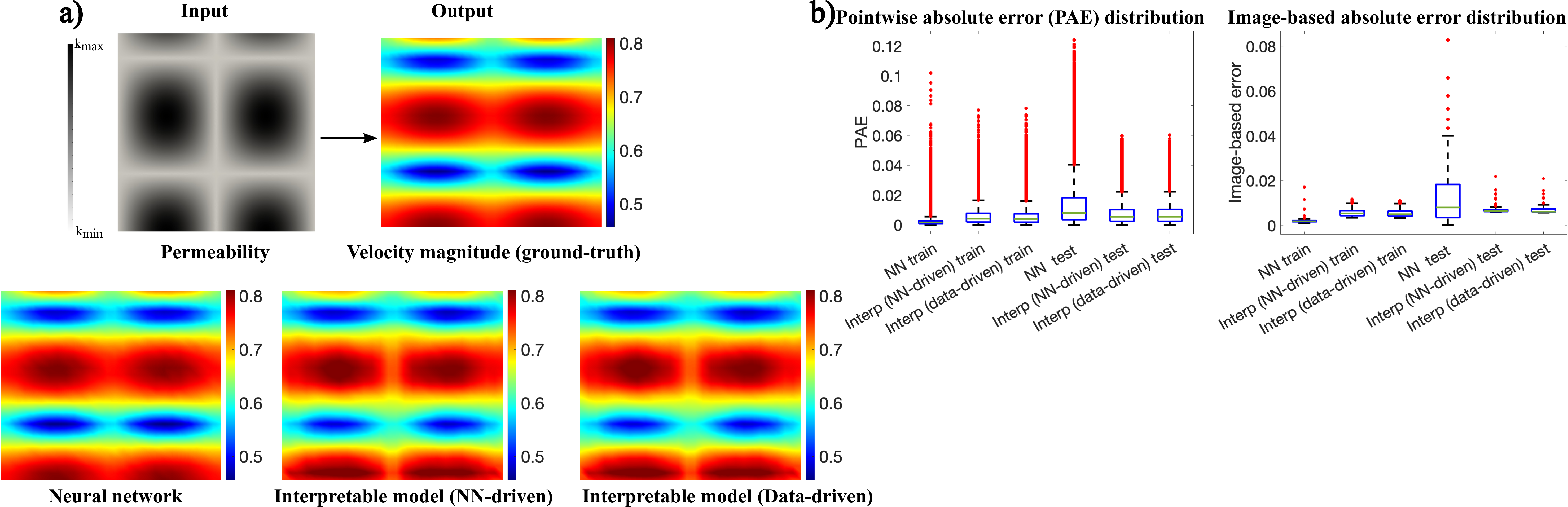}
\caption{Test case 3 results are shown (predicting velocity field from permeability fields in porous media flow). a) An overview of the proposed machine learning task is shown where a 2D velocity magnitude field is predicted from a 2D image (permeability). Neural network, interpretable model trained based on the neural network (NN-driven), and interpretable model trained based on training data (data-driven) results are compared to ground-truth for a sample input in the training regime.  b) Boxplots of the point-wise absolute error (PAE)  distribution considering point-wise error data aggregated across all samples \rev{and image-based absolute error considering the spatially averaged error of each output field variable} are shown. The performance of the neural network (NN), interpretable model trained on neural network predictions (Interp NN-driven), interpretable model trained on training data  (Interp data-driven) are shown for the training data and out-of-distribution test data. Refer to Fig.~\ref{fig:t1} for boxplot details.    } 
\label{fig:t3}
\end{figure}

The same boundary conditions and setup as test case 2 is considered again (without the Brinkman diffusion term). In this test case, more complex permeability patterns are considered and the goal is to predict the 2D velocity magnitude field (image to image mapping). The input permeability field is defined as $k(x,y) = \exp (-4Ax) \lvert \sin(2\pi x)\cos(2\pi B y) \rvert  + 1 $, and $0 \leq A \leq 1 $,   $0 \leq B \leq 4 $ were used in generating 225 simulations used for training. The data were randomly split into 80\% training and 20\% validation.  The goal was to predict velocity magnitude field $\| \bv (x,y) \|$ given $k(x,y)$ as the input function. Out-of-distribution test data were also generated by running 64 simulations within $1 \leq A \leq 2 $ and  $4.2 \leq B \leq 6 $.  In this case, a fully-connected deep autoencoder \edit{with ReLU activation functions} was used. The encoder mapped the input 28$\times$ 28 field to a latent size of 32 through 4 layers \edit{(28$\times$ 28--256--128--64-32),} which was subsequently mapped back to another 28$\times$ 28 field by the decoder with a similar structure as the encoder.  2000 epochs with a learning rate of 5$\times$ 10$^{-4}$, \edit{Adam optimization,} and a batchsize of 64 were used. 

The results are shown in Fig.~\ref{fig:t3}.  The contour plots and the error boxplot show that the neural network makes a better qualitative and quantitative prediction within the training regime. However, similar to the last test cases, the interpretable models have better generalization performance as shown in the boxplot (Fig.~\ref{fig:t3}b) and Table~\ref{tab:mean} and~\ref{tab:max}.

%%%%%%%%%%%%%%%%%%%%%%%%%%%%%%%%%%%%%%%%%%%%%%%%%
\subsection{Test case 4: predicting high-fidelity velocity field from low-fidelity velocity field}
%optimal bandwidth: Matlab book eqn 3.11

\begin{figure}[h!]
\centering
\includegraphics[width=0.9\textwidth]{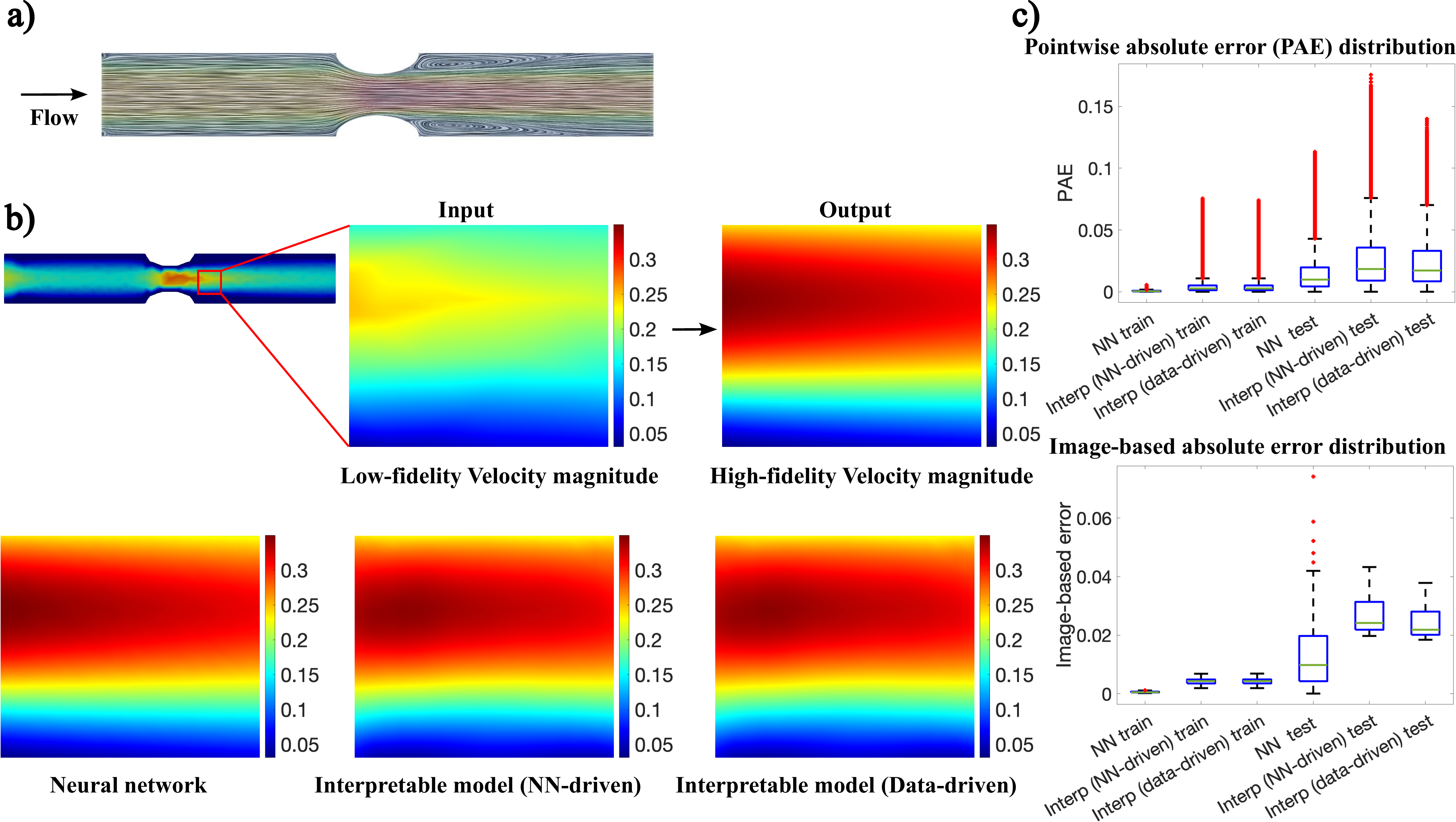}
\caption{Test case 4 results are shown (predicting high-fidelity velocity field from low-fidelity velocity data). a) The simulations are based on steady flow in an idealized blocked vessel. A sample velocity streamline is shown. b) An overview of the proposed machine learning task is shown where high-fidelity 2D velocity magnitude field is predicted from a 2D low-fidelity simulation in the same region of interest. Neural network, interpretable model trained based on the neural network (NN-driven), and interpretable model trained based on training data (data-driven) results are compared to ground-truth for a sample input in the training regime.  c) Boxplots of the point-wise absolute error (PAE)  distribution considering point-wise error data aggregated across all samples \rev{and image-based absolute error considering the spatially averaged error of each output field variable} are shown. The performance of the neural network (NN), interpretable model trained on neural network predictions (Interp NN-driven), interpretable model trained on training data (Interp data-driven) are shown for the training data and out-of-distribution test data. Refer to Fig.~\ref{fig:t1} for boxplot details.     } 
\label{fig:t4}
\end{figure}

An idealized 2D constricted vessel mimicking blood flow in a stenosed artery was considered similar to our prior work~\cite{ArzaniWangDsouza21,Aliakbarietal23} as shown in Fig.~\ref{fig:t4}.  Steady incompressible Navier-Stokes equations were solved for a Newtonian fluid in FEniCS. A parabolic velocity profile was imposed at the inlet and no-slip BC was used at the walls. Training data were generated by performing 400 computational fluid dynamics simulations with different flow rates corresponding to  different Reynolds numbers (defined based on average velocity at the inlet) between 15 and 225. In the high-resolution finite element simulations, quadratic and linear shape functions were used for velocity and pressure, respectively (P2-P1 elements) with 41.4k triangular elements. Similarly, low-resolution (low-fidelity) simulations were performed by increasing the viscosity by 20\% (representing a dissipative solution with artificial diffusion) and using first order velocity elements (P1-P1 elements) with a total of 536 elements. The goal of the machine learning models is to predict the high-fidelity velocity magnitude field $\| \bv_{hres} (x,y) \|$ from the low-fidelity field $\| \bv_{lres} (x,y) \|$. We focus on a specific region of interest downstream of the stenosis as shown in  Fig.~\ref{fig:t4}b. Superresolution with machine learning is an active area of research in fluid mechanics~\cite{fukami2023super}, and additionally, prior machine learning models have dealt with mapping between multi-fidelity data~\cite{de2022neural,aliakbari2022predicting}. In our example, both datasets are first interpolated to a structured 28$\times$28 grid. 100 out-of-distribution high-resolution and low-resolution simulations were also performed by varying the Reynolds number between 240 and 300. The neural network architecture was a deep autoencoder similar to test case 3 but with one additional encoder and decoder hidden layer \edit{(the encoder architecture was 28$\times$ 28--512--256--128--64-32 and the decoder was its symmetric counterpart.)} The training data were randomly split into 80\% training and 20\% validation. 5000 epochs with a learning rate of 2.5$\times$ 10$^{-5}$ and  a batchsize of 64 were used. Finally, in this test case, instead of using a broad range for the candidate bandwidths in the interpretable model (Table~\ref{tab:cand}), we select a focused range estimated based on existing plug-in methods for optimal bandwidth selection. Namely, $\beta_{opt} = \mathcal{O}( n^{-0.3}) $ has been proposed as an optimal bandwidth for Gaussian kernels~\cite{horova2012kernel,altman1995bandwidth}. Considering n=28 as the number of points in each direction, $\beta_{opt} \approx 0.37 $. Therefore, we focused on $0.2<\beta<0.4$ in constructing our library (Table~\ref{tab:cand}). We verified that this range gave optimal training errors compared to other choices. It should be noted that the problem of optimal bandwidth selection is complicated~\cite{horova2012kernel,kohler2014review}, particularly for our problem where different kinds of kernels and generalized linear models are used. 

The contour plots and the error boxplots are shown in Fig.~\ref{fig:t4}. The neural network produces very accurate training results indistinguishable from the ground-truth. The interpretable model results also mimic the key quantitative and qualitative patterns with minor distinctions visible. In this test case, the interpretable models could not improve the \rev{average} out-of-distribution test errors compared to the neural network \rev{and only reduced the maximum image-based absolute error.}

%%%%%%%%%%%%%%%%%%%%%%%%%%%%%%%%%%%%%%%%%%%%%%%%%
\subsection{Test case 5: predicting high-fidelity wall shear stress field from low-fidelity velocity data away from the wall}

\begin{figure}[h!]
\centering
\includegraphics[width=0.9\textwidth]{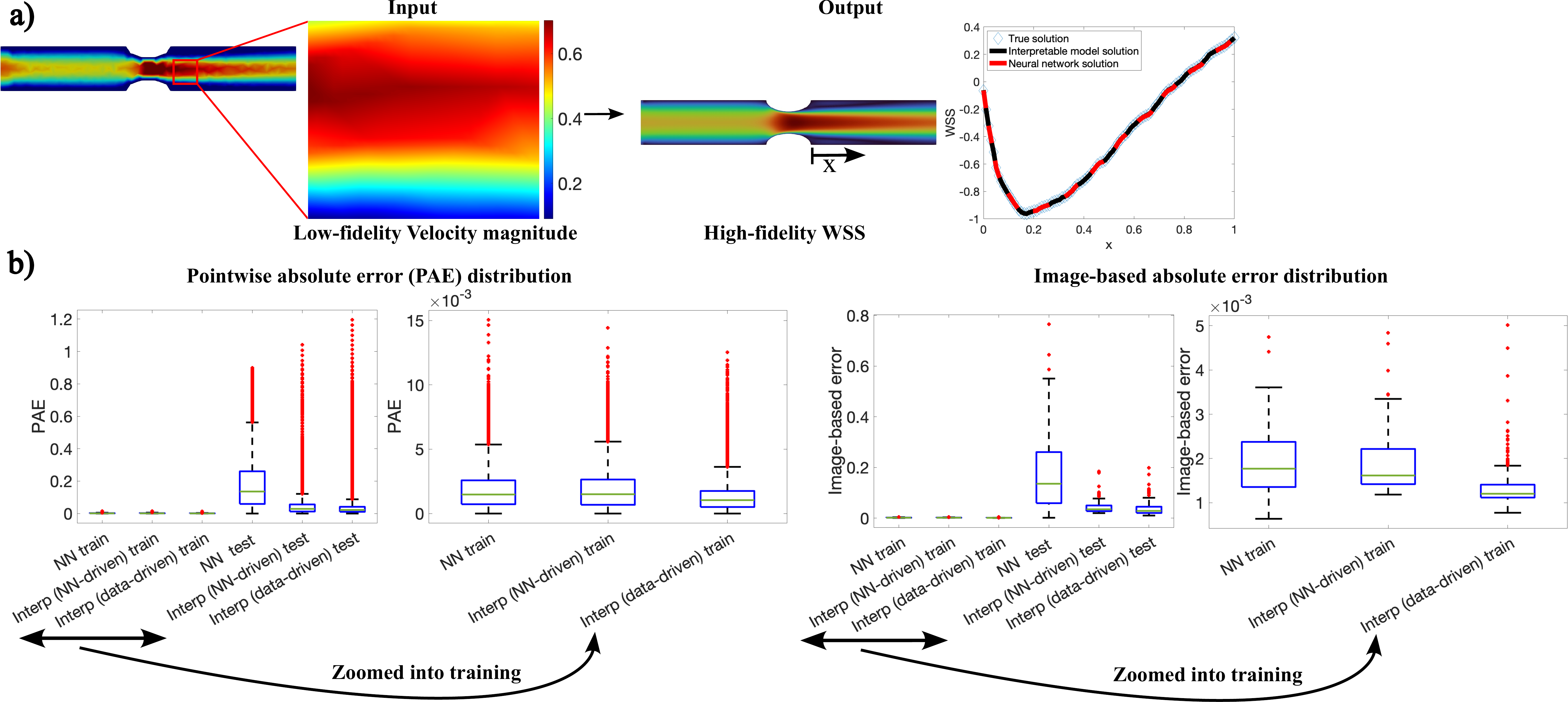}
\caption{Test case 5 results are shown (predicting wall shear stress WSS field from low-fidelity velocity data). The same model as test case 4 is used. a) An overview of the proposed machine learning task is shown where high-fidelity WSS field is predicted from a 2D low-fidelity simulation. Neural network, interpretable model trained based on the neural network (NN-driven), and interpretable model trained based on training data (data-driven) results are compared to ground-truth for a sample input in the training regime as shown in the WSS vs.\ x plot.  b) Boxplots of the point-wise absolute error (PAE)  distribution considering point-wise error data aggregated across all samples \rev{and image-based absolute error considering the spatially averaged error of each output field variable} are shown. The performance of the neural network (NN), interpretable model trained on neural network predictions (Interp NN-driven), interpretable model trained on training data  (Interp data-driven) are shown for the training data and out-of-distribution test data. Refer to Fig.~\ref{fig:t1} for boxplot details.    } 
\label{fig:t5}
\end{figure}

In this example, we reconsider the exact same dataset in the constricted artery model of the previous test case. The goal of the machine learning model here is to take the low-fidelity velocity magnitude field in the same region of interest (away from the wall) and predict high-fidelity wall shear stress (WSS) at the bottom wall as shown in Fig.~\ref{fig:t5}. In this case, the machine learning model needs to map a 2D scalar field to a 1D scalar field. A deep autoencoder similar to test case 3 was used with the last encoder layer being mapped to a 100 $\times$ 1 line instead of an image. 5000 epochs with a learning rate of 2.5$\times$ 10$^{-5}$ and  a 64 batchsize were used. 

As shown in  Fig.~\ref{fig:t5}, all methods provide a very accurate estimate for WSS in the training regime. In this case, the distinction between the training and test errors was more pronounced for both neural network and interpretable models. As seen more clearly in Table~\ref{tab:mean} and~\ref{tab:max}, in testing, the mean absolute error was considerably reduced for the interpretable models. Another interesting observation was that the data-driven interpretable model had \rev{slightly} better training performance compared to the neural network model.

%%%%%%%%%%%%%%%%%%%%%%%%%%%%%%%%%%%%%%%%%%%%%%%%%
\subsection{Test case 6: local explanation of neural network predictions in a porous media flow example}

\begin{figure}[h!]
\centering
\includegraphics[width=0.9\textwidth]{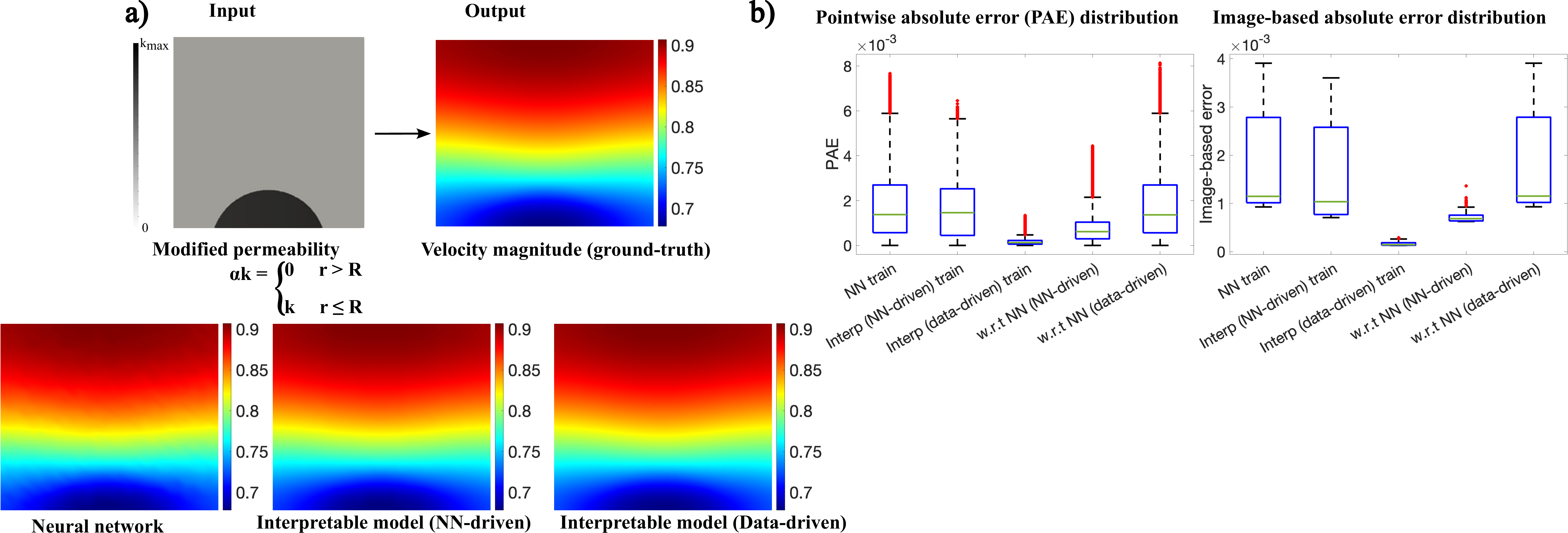}
\caption{Test case 6 results are shown (predicting velocity field from permeability fields in porous media flow by locally probing the neural network). a) An overview of the proposed machine learning task is shown where a 2D velocity magnitude field is predicted from a 2D image (permeability). Neural network, interpretable model trained based on the neural network (NN-driven), and interpretable model trained based on training data (data-driven) results are compared to ground-truth for a sample input in the training regime.  b) Boxplots of the point-wise absolute error (PAE)  distribution considering point-wise error data aggregated across all samples \rev{and image-based absolute error considering the spatially averaged error of each output field variable} are shown. The performance of the neural network (NN), interpretable model trained on neural network predictions (Interp NN-driven), interpretable model trained on training data  (Interp data-driven) are shown for the training data and out-of-distribution test data.  Interpretable model performance with respect to the locally probed NN is also shown in this test case. Refer to Fig.~\ref{fig:t1} for boxplot details.      } 
\label{fig:t6}
\end{figure}

In all of the previous test cases, we used the exact same data used in training the neural network to train the proposed interpretable models. However, this is not required for the NN-driven Interp model. Namely, the trained neural network could be probed for any desired input to generate pairs of input-output data for training the NN-driven Interp model. In the case where one is interested in explaining the neural network behavior within the training regime, the NN-driven Interp model will be trained with a combination of training and in-distribution test data. 

In this last test case, we consider the porous media flow in test case 2. We reconsider the problem where the goal is to predict the velocity magnitude (instead of maximum velocity) from the input modified permeability field as shown in Fig.~\ref{fig:t6}. The same dataset used in test case 2 is used for training the neural network. A \edit{fully connected} autoencoder \edit{with ReLU activation functions} mapped the input 28$\times$ 28 field to a latent size of 8 through 4 layers \edit{(28$\times$ 28--256--128--64-8),} which was subsequently mapped back to another image by a similar decoder.  2000 epochs with a learning rate of 5$\times$ 10$^{-4}$ and  a batchsize of 64 were used. The neural network was trained on the entire dataset explained in test case 2. However, the goal here was to interpret the neural network predictions locally. The position of the porous region was fixed at R=0.02 and Y=-0.1. The trained network was probed for 100 different A values (permeabilities) ranging between $0 \leq A \leq 2 $. This represented a local probing of the neural network with a higher sampling rate than what was used for its training. Finite element simulations were also performed for error quantification. 

The results are shown in  Fig.~\ref{fig:t6}. A data-driven Interp model was also trained based on the ground-truth data for comparison. The NN-driven Interp model produced very accurate results and could faithfully explain the neural network behavior in this localized region of the training landscape. An interesting observation is that the NN-driven Interp model slightly improves the training error compared to the neural network model and produces slightly smoother qualitative patterns. The data-driven Interp model produces significantly more accurate results compared to the neural network model. This should not be surprising because in this case the data-driven Interp model was trained based on the ground-truth data in a localized parameter space, whereas the neural network was trained over a larger parameter space. \edit{In other words, it is not fair to compare the data-driven Interp results to the neural network in this case.} Test errors are not shown in Fig.~\ref{fig:t6}b as in this case the Interp models were not trained based on the entire data. Instead, the errors in interpretable model predictions with respect to the neural network predictions are shown. As expected, the NN-driven Interp case matches the NN behavior more closely compared to the data-driven Interp case. The difference between the two interpretable models was less in most previous test cases where global interpretation instead of local interpretation was done.

%%%%%%%%%%%%%%%%%%%%%%%%%%%%%%%%%%%%%%%
\section{Discussion} \label{sec:disc}

In this study, we proposed an interpretable surrogate model that approximates neural network's predictions locally or globally.  The interpretable model was in the form of integral equations inspired by  functional linear models. We applied our framework to different deep learning models trained on making predictions based on functions and functionals in different physics-based problems. The results demonstrated that in most test cases the interpretable model improved generalization error and even in some cases training error was improved compared to the neural network. Our proposed approach for improving generalization error could be compared to the process of human thinking. When we are asked questions that are outside our knowledge domain we probe the existing knowledge in our brain and we generate an answer to the new questions by using interpretation and reasoning. The proposed NN-driven interpretable model could be perceived within this context where we probe the neural network (our existing knowledge) to build an interpretable model to answer an unknown question (an OOD input).

A surprising observation was the improved training error in the interpretable model compared to the deep learning model in some cases. In test case 1 (EMNIST), the mean training errors were reduced by NN-driven and data-driven interpretable models, and in test case 5 the data-driven interpretable model reduced the mean and \rev{point-wise} peak training errors. Also, in some other cases (e.g., test case 3), the maximum training error was reduced. \rev{In-distribution generalization (validation) results shown in the Appendix (Sec~\ref{sec:app_valid}) demonstrated further improvements in the interpretable model performance compared to the deep learning counterpart.}  Training error improvement by the NN-driven interpretable model observed in certain cases was a particularly unprecedented result that could be attributed to the smoothing effect in functional linear models, which has been well studied in the context of kernel smoothing~\cite{horova2012kernel,ghosh2018kernel}.  Except for test case 4, the interpretable models consistently exhibited reduced test error across all cases. This suggests that interpretable models have the potential to enhance predictive accuracy and generalize well to unseen data, showcasing their effectiveness in improving model performance.

A notable characteristic of our proposed framework is its inherent flexibility. Our interpretable model could be built either based on the neural network predictions (NN-driven) or the training data without the need for a neural network (data-driven).  The former is preferred when an interpretation of an \edit{opaque} neural network model is desired, while the latter is preferred where improved accuracy (particularly improved OOD generalization) is desired. Our framework also shares many of the advantages offered by other operator learning models. For instance, similar to neural operators our framework once trained could be used to evaluate the solution at any desired input location, rather than being restricted to fixed locations as in traditional neural networks~\cite{kovachki2023neural}. It has been shown in prior operator learning work with DeepONets that a small amount of data can improve their generalization error~\cite{zhu2023reliable}. It has also been demonstrated that sparsity promoting neural network architectures can have good performance with small training data~\cite{lemhadri2021lassonet,de2022neural}. Our proposed interpretable model promotes a sparse solution to the operator learning problem, and therefore even just a small amount of OOD training data is expected to even further improve its OOD generalization, which should be investigated in future work. 

In related work, deep learning has been used to discover extensions of Green's functions beyond linear operators~\cite{gin2021deepgreen,boulle2022data}. It is known that approximating Green's functions with neural networks is easier than approximating the action of Green's function on the input (Green's operator)~\cite{boulle2022data}.  This is consistent with our framework where we learn kernel functions in our integral equations. Another analogy could be made with Koopman operators, which provide a theoretical framework for linearizing dynamical systems~\cite{budivsic2012applied,mezic2013analysis} and have been approximated with \edit{opaque} neural networks~\cite{takeishi2017learning}. Dynamic mode decomposition (DMD) is an interpretable numerical approximation of the Koopman operator. DMD's interpretability is improved by retaining fewer modes or using sparsity promoting approaches~\cite{jovanovic2014sparsity}. This is similar to our framework where an interpretable model is selected in the form of generalized functional linear models to approximate an unknown operator. Additionally, the tradeoff between accuracy and interpretability is similar where reducing the number of modes in DMD (or the number of integral equations in our framework) increases interpretability at the cost of potentially reduced accuracy. \edit{ Our surrogate model could be perceived as a reduced-order-model (ROM) that approximates the neural network behavior and as such, just like how ROMs can simplify understanding of a complex system, our model can be used towards a similar goal (which remains to be investigated). Additionally, each integral term is equipped with a coefficient that tells the significance of the term. Therefore, once we identify the significance of each term, we can understand the network based on its kernel and associated bandwidth. For example, if a kernel with a large  bandwidth is important in the total response, then long-range effects in the input image affect the output. Similarly, in DMD, each mode comes with a frequency that provides information about the dynamics of the system.  }

The utilization of a library of candidate models has been leveraged in other scientific machine learning problems. Sparse identification of nonlinear dynamics (SINDy) models a nonlinear dynamical system by constructing analytical equations in the form of a nonlinear system of ordinary differential equations, where the terms in the equations are selected from a pre-specified library~\cite{brunton2016discovering}. As another example, a library of hyperelastic constitutive equations has been used for discovering constitutive models in nonlinear solid mechanics problems~\cite{flaschel2021unsupervised}. Machine learning ROMs have been proposed where a library of proper orthogonal decomposition (POD) modes are used for parameter identification from low-resolution measurement data~\cite{BrightLinKutz13,ArzaniDawson21}. Another analogy can be drawn with ensemble machine learning models. Neural additive models use an ensemble of parallel neural networks and make final predictions with linear superposition~\cite{agarwal2021neural}. Similarly, our approach could be perceived as an ensemble of approximations to the solution (each integral equation) that is linearly added to build the final solution. 

Our proposed framework offers the flexibility to be extended to other deep learning tasks. For instance, in certain tasks in addition to a field variable, some physical parameters might also be inputs to the neural network. As an example of an extension to such cases, the scalar response model (Eq.~\ref{eqn:fda2}) could be extended as $\mathbf{u} = r(z) \int  \bm{\psi}(\bm{\xi}) \mathbf{f}(\bm{\xi})   \,d \bm{\xi}  + \gamma z $ similar to the work in~\cite{li2010generalized} where $z$ is the additional input parameter, and $r$ and $\gamma$ are an unknown function and parameter, respectively, that need to be estimated. Leveraging analytical integral equation models in classical physics is another possible extension.  An example of analytical integral equations used in fluid dynamics is the Biot-Savart Law used in modeling vortex dynamics~\cite{panton2013incompressible}. This has recently inspired the neural vortex methods, which use neural networks to map vorticity to velocity~\cite{xiong2023neural}. Our analytical integral equation approach also offers the possibility of solving inverse problems using standard approaches used in solving integral equations~\cite{nair2011advanced}. Integral equations have been utilized in developing mathematical theories for inverse problems and their numerical solution~\cite{isakov2006inverse,delillo2003detection}. Another interesting future direction is the comparison of our method's generalization with other operator learning methods such as DeepONets~\cite{lu2021learning} and Fourier neural operators~\cite{li2020fourier}. Extension to time-dependent problems is another future direction, which is inspired by  parabolic Green's functions~\cite{boulle2022learning}. Finally, our definition of interpretability draws from qualitative attributes outlined in~\cite{sudjianto2021designing} such as additivity, sparsity, and linearity, as well as being able to present the model as an analytical equation. \edit{Our current work just focused on demonstrating the possibility of approximating neural networks with analytical models that possess such interpretable features and we did not demonstrate our framework's potential for physical interpretation.}  Our future work will focus on using the model for interpreting the physics of the problem. 

\section{Conclusion} \label{sec:conc}

We have proposed an interpretable surrogate model to not only interpret a given neural network but also improve generalization and extrapolation. Our results demonstrate very good and comparable training error and in most cases improved OOD generalization error once compared to the neural network. In a broader sense,  our framework suggests the notion of a hybrid machine learning strategy where a trained deep learning model is used for in-distribution predictions and an interpretable surrogate is utilized for OOD predictions.  This hybrid strategy could be compared with hybrid finite-element and neural network strategies recently proposed to improve neural network predictions~\cite{liang2023synergistic}. Our study suggests that by leveraging integral equations in the form of generalized functional linear models, we can build more interpretable and explainable scientific machine learning models with a high potential for improved generalization.

%%%%%%%%%%%%%%%%%%%%%%%%%%
\section*{Acknowledgement}
This work was supported by NSF Award No.~2247173 from NSF's Office of Advanced Cyberinfrastructure (OAC). Additionally, support from NSF IIS-2205418 and DMS-2134223 is acknowledged. We would like to thank Dr. Emma Lejeune and Dr. Harold Park for discussions related to this work and assistance in using the MNIST/EMNIST datasets. \rev{We are also grateful to the anonymous Referee for their constructive comments that improved the paper.} 

\section*{Competing Interests}
The authors have no conflicts of interest.

\section*{Data Availability}
\edit{The codes and data used to generate the results in the manuscript can be accessed at \url{https://github.com/amir-cardiolab/XAI_FDA}. 
}

\section{Appendix} 
\subsection{Normal equations for functional linear models} \label{sec:app}
Here, we present an alternative strategy for finding the kernels in functional linear models using the normal equations, based on the presentation in~\cite{horvath2012inference}. Let's consider the fully functional model, which was used for image to image mapping in this study (Eq.~\ref{eqn:fda1}) in the scalar form

 \begin{equation} 
\mathbf{u}(\bx) = \int  \psi( \bm{\xi}, \bx) \bm f(\bm{\xi})   \,d \bm{\xi}   \;,
\label{eqn:fda11}
\end{equation}
where given $Q$ pairs of training data, we have grouped them as column vectors  $\bm u(\bx) = \left[ u_1(\bx) , \dots,  u_Q(\bx)  \right]^T  $ and  $\bm f(\bm\xi) = \left[ f_1(\bm\xi) , \dots,  f_Q(\bm\xi)  \right]^T  $. We expand the unknown kernel function in Eq.~\ref{eqn:fda11} using pre-defined arbitrary bases as
\begin{equation} 
\psi( \bm{\xi}, \bx)   = \sum_{i} \sum_{j}   b_{ij}  \omega_i(\bm\xi) \theta_j(\bx)    \;, 
\end{equation}
where $\omega_i$ and  $\theta_j$ are the basses and $b_{ij}$ are the unknown coefficients that could be grouped into a matrix $\mathbf{B} = \left[ b_{ij} \right] $. Our goal is to solve the following least squares problem
\begin{equation} 
\min_{ \bm{\psi}}  \sum_{n=1}^{Q} \|     u_n (\bx) - \int  \psi( \bm{\xi}, \bx) f_n(\bm{\xi})   \,d \bm{\xi}        \| ^2     \;.
\end{equation}
Grouping the bases into column vectors $\bm\omega (\bm{\xi}) = \left[\omega_1(\bm{\xi}) , \dots \right]^T  $ and $\bm\theta(\bx) = \left[\theta_1(\bx) , \dots \right]^T  $, we can rewrite Eq.~\ref{eqn:fda11} in matrix form as

\begin{equation} 
\mathbf{u} (\bx) = \mathbf{Z} \mathbf{B} \bm\theta(\bx)  \;,
\end{equation}
where $ \mathbf{Z} = \int  \mathbf{f}(\bm{\xi}) \bm\omega^T(\bm\xi)   \,d \bm{\xi}   $. Finally, by defining the matrix $\mathbf{J} = \int \bm\theta(\bx)  \bm\theta^T(\bx)    \,d \bx   $, we can derive the final form of the normal equations

\begin{equation} 
 \mathbf{Z}^T  \mathbf{Z}   \mathbf{B} \mathbf{J} =  \mathbf{Z}^T  \int  \bm u(\bx)    \bm\theta^T(\bx)    \,d \bx     \;,
\end{equation}
where we need to solve for $ \mathbf{B}$. 

We can also write a similar version of the above equation by reconsidering the optimization problem in Eq.~\ref{eqn:sindy}, which was used for approximating the solution of $\mathbf{U} = \mathbf{F}\mathbf{W}$ in Sec.~\ref{sec:flm}.  Instead of introducing an L1-regularized problem as done in Eq.~\ref{eqn:sindy}, we can directly solve this regression problem using the normal equations

\begin{equation} 
\mathbf{F}^T \mathbf{F} \mathbf{W}  =  \mathbf{F}^T \mathbf{U} \;.
\end{equation}
This equation could be solved using a linear solver to find $\mathbf{W}  $. However, in practice the $\mathbf{F}^T \mathbf{F}$ matrix is highly ill-conditioned and close to singular, therefore an L2 regularization should be added 
\begin{equation} 
( \mathbf{F}^T \mathbf{F} + \lambda \mathbf{I} ) \mathbf{W}  =  \mathbf{F}^T \mathbf{U} \;,
\end{equation}
where $\lambda$ is the regularization parameter. An increased $\lambda$  provides a more robust linear system of equations but at the cost of reduced accuracy.  Our preliminary investigation has shown that this formulation in certain cases produces more accurate results related to the training error. The OOD generalization error was better in most cases for the L1-regularized problem (except for test case 2). It should also be noted that the L2-regularized problem produces a dense solution where most integral equations in the library will be nonzero, and therefore a less interpretable model is produced. 

%Inference FDA book Sec 8.3
\rev{
\subsection{In-distribution generalization} \label{sec:app_valid}

In this section, we present the in-distribution generalization (validation) errors for all the test cases where global interpretation was performed (cases 1--5). The simulations used in evaluating the validation errors were sampled from the same parametric space defined in the problems but different from the training data to assess the in-distribution generalization (interpolation) accuracy of the models. The validation datasets used in assessing the deep learning and XAI models in this section are identical. The point-wise absolute error and image-based absolute error distributions for the validation data are shown in Fig.~\ref{fig:val}. Additionally, Table~\ref{tab:val_mean} and~\ref{tab:val_max} present the mean and maximum validation errors. Comparing the validation errors with the previously presented training errors demonstrates reasonable performance for all models. Interestingly, in some cases, the interpretable models have slightly better validation errors compared to the neural networks. Additionally, in some cases, the validation errors are slightly better than the training error, which is because the most challenging data (e.g., higher Reynolds number) is included in the training set. 

\begin{figure}[h!]

\centering
\includegraphics[width=\textwidth]{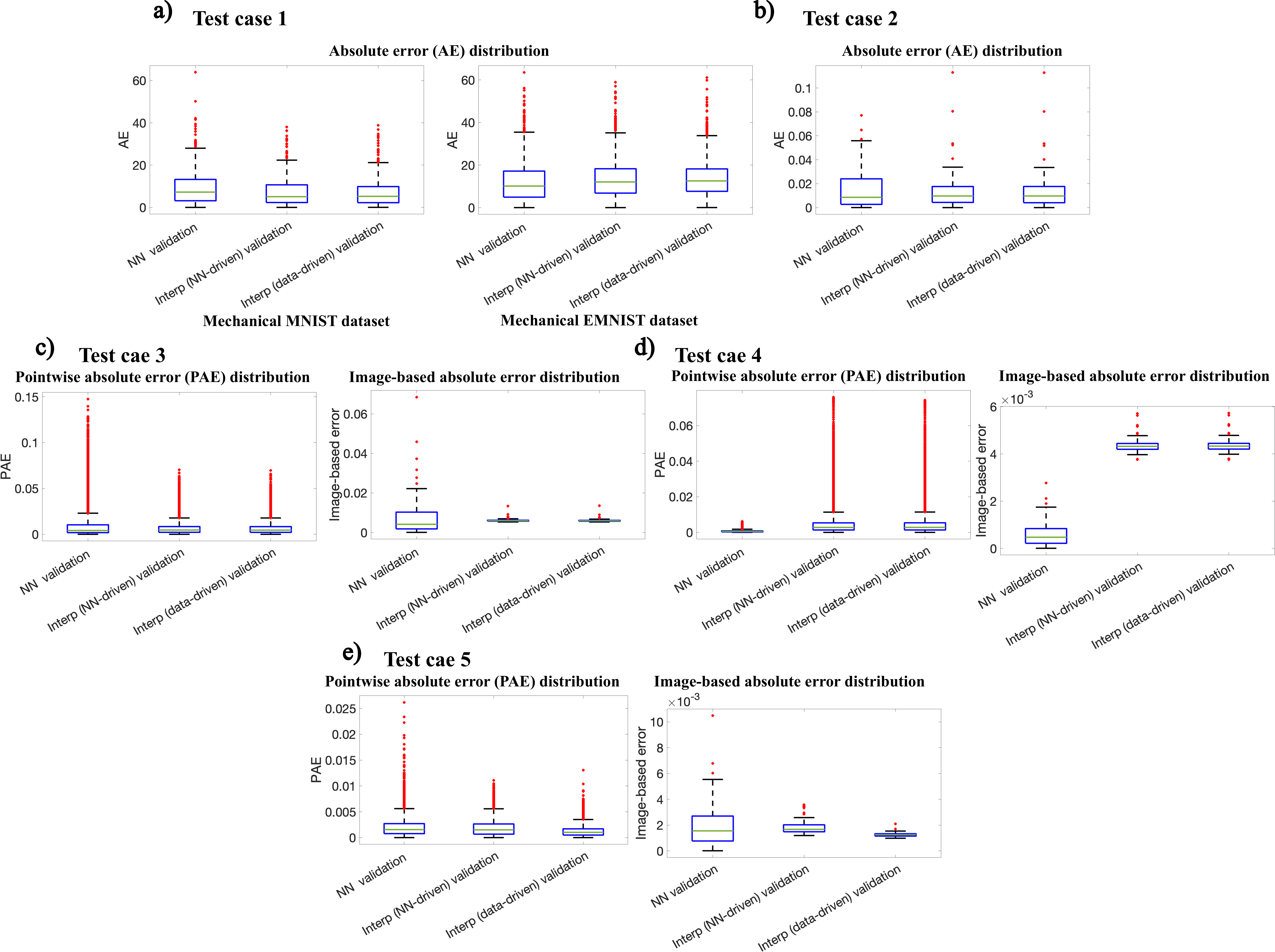}
\caption{\rev{ In-distribution generalization (validation) error distributions are plotted for the first five test cases. In test cases with mapping to field variables, point-wise absolute error (PAE) aggregates all of the point-wise errors, whereas image-based error calculates the spatially averaged error of each output field variable.    } } 
\label{fig:val}

\end{figure}

\begin{table}[h!]
\centering
\begin{small}
\rev{
\begin{tabular}{ |p{1.8cm}||p{1.4cm}|p{1.8cm}|p{2.1cm}|  }
 \hline
 \multicolumn{4}{|c|}{Mean absolute error (MAE) for the validation data} \\
 \hline
Test case& NN &  Interp NN-driven & Interp data-driven\\
 \hline
case~1 \begin{footnotesize}(MNIST)\end{footnotesize}   & 9.49& 7.63  &   7.31    \\
case~1 \; \begin{footnotesize}(EMNIST)\end{footnotesize}  & 12.23 & 13.33&   13.64     \\
case 2 & 0.015 & 0.014&   0.014  \\
case 3    & 0.0087  & 0.0062 &   0.0061  \\
case 4    & 0.0006  &  0.0044&   0.0044   \\
case 5 &  0.002  & 0.0018&   0.0013 \\
 \hline
\end{tabular}
}
 \end{small}
\caption{\rev{Mean absolute error (MAE) for the neural network (NN), interpretable model trained on neural network predictions (Interp NN-driven), and interpretable model trained on training data (Interp data-driven) are listed for in-distribution testing (validation) data. Test case 6 was based on local interpretation and is not included.}   }
\label{tab:val_mean}
\end{table}

\begin{table}[h!]
\centering
\begin{small}
\rev{
\begin{tabular}{ |p{3.3cm}||p{1.4cm}|p{1.8cm}|p{2.1cm}|  }
 \hline
 \multicolumn{4}{|c|}{Maximum absolute error for the validation data} \\
 \hline
Test case& NN &  Interp NN-driven & Interp data-driven\\
 \hline
case~1 \begin{footnotesize}(MNIST)\end{footnotesize}   & 63.87 & 38.03  &   38.82    \\
case~1 \; \begin{footnotesize}(EMNIST)\end{footnotesize}  &  63.48 & 58.97 &   61.15   \\
case 2 & 0.077 & 0.11&   0.11  \\
case 3 PAE    & 0.15  & 0.070 &   0.070  \\
case 3 image-based    & 0.068  & 0.013 &   0.014  \\
case 4 PAE    & 0.0062  & 0.076 &   0.074  \\
case 4 image-based    & 0.0028  & 0.0057 &   0.0057  \\
case 5 PAE    & 0.026  & 0.011 &   0.013  \\
case 5 image-based    & 0.010  & 0.0036 &   0.0021  \\
 \hline
\end{tabular}
}
 \end{small}
\caption{\rev{Maximum absolute error for the neural network (NN), interpretable model trained on neural network predictions (Interp NN-driven), and interpretable model trained on training data (Interp data-driven) are listed for in-distribution testing (validation) data. In cases where the output is a field, maximum error is either calculated based on point-wise data aggregated across all samples (PAE) or in an image-based fashion as the spatially averaged error of each output field variable.  Test case 6 was based on local interpretation and is not included.}   }
\label{tab:val_max}
\end{table}

\subsection{Error percentiles} \label{sec:app_perc}

Given the sensitivity of the maximum point-wise absolute error (PAE) to outliers due to its point-wise nature, Table~\ref{tab:perc} presents the PAE percentiles for the training and OOD test data.  PAE percentiles for the validation data are listed in Table~\ref{tab:perc2}. In general, the tables indicate that the trends in comparisons remain consistent.

\begin{table}[h!]
\centering
\begin{small}
\rev{
\begin{tabular}{ |p{1.8cm}||p{1.8cm}|p{1.8cm}|p{2.5cm}|p{1.7cm}|p{1.8cm}|p{2cm}| }
 \hline
 \multicolumn{7}{|c|}{Error percentiles (Perc) for point-wise absolute error (training and test data).} \\
 \hline
Test case& NN (train) &  Interp NN-driven (train) & Interp data-driven (train) & NN (test) & Interp NN-driven (test) & Interp data-driven (test)\\
 \hline
case~1 \begin{footnotesize}(MNIST)\end{footnotesize}   & &   &     &    & &     \\
99th Perc & 27.2 & 33.33  & 33.55     & 105.46    & 41.53 & 24.43     \\
97th Perc & 17.23 &24.62   & 24.44      & 87.25    & 37.38 & 21.07     \\
95th Perc & 13.41 & 21.43  & 20.81     & 75.57    & 34.69 & 19.42     \\
case~1 \; \begin{footnotesize}(EMNIST)\end{footnotesize} & &   &     &    & &   \\
99th Perc & 34.54 & 35.63   &  33.94    &  232.87   & 227.44 & 174.79     \\
97th Perc.& 28.08 & 27.66   &  26.48    & 216.25    &  205.98 &  159.41    \\
95th Perc & 24.98 & 24.39   & 23.13     & 204.94    & 193.27 & 148.81     \\
case 2 & &   &     &    & &  \\
99th Perc & 0.014  & 0.050   & 0.049     & 0.37    & 0.17  & 0.18     \\
97th Perc & 0.011  & 0.031  &  0.031   &  0.36  & 0.17 & 0.17    \\
95th Perc & 0.01 & 0.027  & 0.027     &  0.33  & 0.16 & 0.16     \\
case 3    & &   &     &    & & \\
99th Perc &0.0082  & 0.026   & 0.025     & 0.077   & 0.032  & 0.031     \\
97th Perc& 0.0058 & 0.019   &  0.019    &  0.062   & 0.023 &  0.023    \\
95th Perc & 0.005 & 0.016  & 0.016    &   0.054   & 0.020 & 0.020     \\
case 4    & &   &     &    & &  \\
99th Perc & 0.0022  & 0.026  & 0.026     & 0.064   & 0.11 & 0.10     \\
97th Perc & 0.0018 &  0.016  &  0.016    & 0.050    & 0.092  & 0.080    \\
95th Perc & 0.0015 & 0.013  &  0.013    & 0.043    & 0.080 & 0.070    \\
case 5& &   &     &    & & \\
99th Perc & 0.0066  & 0.0067  & 0.0055     &  0.69  & 0.36 & 0.47     \\
97th Perc& 0.0054 & 0.0054  & 0.0042     &  0.57   & 0.16 & 0.20    \\
95th Perc &  0.0048 & 0.0048   &  0.0035     &  0.51   & 0.13 & 0.12     \\
case 6 & &   &     &    & & \\
99th Perc & 0.0056 & 0.0051  & 0.00075     &--    & -- & --    \\
97th Perc &  0.005 & 0.0044   & 0.0006   & --   &-- &   --  \\
95th Perc & 0.0045 & 0.004  &  0.0005    & --   &-- &--     \\
 \hline
\end{tabular}
}
 \end{small}
\caption{\rev{Error percentiles for the neural network (NN), interpretable model trained on neural network predictions (Interp NN-driven), and interpretable model trained on training data (Interp data-driven) are listed for training and out-of-distribution testing. Test case 6 was based on local interpretation and did not evaluate test data. Additionally, the errors reported for test case 6 are based on the local data used for local evaluation. } }
\label{tab:perc}
\end{table}

\begin{table}[h!]
\centering
\begin{small}
\rev{
\begin{tabular}{ |p{3.3cm}||p{1.4cm}|p{1.8cm}|p{2.1cm}|  }
 \hline
 \multicolumn{4}{|c|}{Error percentiles (Perc) for point-wise absolute error (validation data)} \\
 \hline
Test case& NN &  Interp NN-driven & Interp data-driven\\
 \hline
case~1 \begin{footnotesize}(MNIST)\end{footnotesize}   &  &  &      \\
99th Perc  & 39.39  & 32.32 & 33.74 \\
97th Perc  & 29.95 & 26.05 & 26.09 \\ 
95th Perc  & 27.2  &  23.44&  21.32\\
case~1 \; \begin{footnotesize}(EMNIST)\end{footnotesize}  &  &  &    \\
99th Perc  & 42.12  & 40.71  & 39.33 \\
97th Perc  &  34.53 & 32.78 & 31.41 \\ 
95th Perc  &  31.12 & 29.56 & 27.96 \\
case 2 &  &  &   \\
99th Perc  &  0.071& 0.097  & 0.097   \\
97th Perc  & 0.057  & 0.053 & 0.053 \\ 
95th Perc  & 0.053 & 0.047 & 0.046 \\
case 3    &  &  &   \\
99th Perc  & 0.059  & 0.029 & 0.029 \\
97th Perc  & 0.041 &0.021 & 0.020 \\ 
95th Perc  & 0.033 & 0.017 & 0.017 \\
case 4   &  &  &   \\
99th Perc  & 0.0023  & 0.027 & 0.027 \\
97th Perc  & 0.0018 & 0.017 & 0.017 \\ 
95th Perc  &  0.0016 & 0.014 & 0.014 \\
case 5     &  &   &    \\
99th Perc  &  0.0074  & 0.0067  & 0.0051  \\
97th Perc  & 0.0058 & 0.0054 & 0.0038 \\ 
95th Perc  & 0.0050  & 0.0048 & 0.0033 \\

 \hline
\end{tabular}
}
 \end{small}
\caption{\rev{Error percentiles for the neural network (NN), interpretable model trained on neural network predictions (Interp NN-driven), and interpretable model trained on training data (Interp data-driven) are listed for the validation data. Test case 6 was based on local interpretation and is not included. }   }
\label{tab:perc2}
\end{table}

\subsection{Statistical significance} \label{sec:app_stat}
Statistical analysis was performed to assess the statistical significance of the differences observed in the error distributions for each test case based on the point-wise data. First, Friedman’s test was performed in a three-way manner considering the NN, Interp NN-driven, and Interp data-driven models. Separate Friedman tests were performed for the training, validation, and OOD test datasets.  Subsequently, after verifying statistical significance,  one-on-one tests were performed using Wilcoxon's signed rank test. This was done for all training, validation, and OOD test datasets, where within each case all possible pairs (e.g., NN vs.\ Interp NN-driven) were tested to ensure the differences in errors are significant. To account for the moderate/large sample size, Good's q-value~\cite{woolley2003p} was used instead of the regular p-value. q-value smaller than 0.005 was considered statistically significant. 

All Friedman q-values were significant with the exception of case 2's validation errors. Therefore, the differences observed in the performance of different methods on the validation dataset for case 2 were not statistically meaningful (training and OOD differences were meaningful for case 2). The subsequent Wilcoxon test on all of the other cases revealed statistically significant results with the exception of case 5's comparison between the NN and Interp NN-driven models on the training dataset (q-value = 0.11 and p-value = 0.0058).

}
\clearpage

\bibliographystyle{unsrt}
\bibliography{bib-interp}

\begin{thebibliography}{10}

\bibitem{fukami2023super}
K.~Fukami, K.~Fukagata, and K.~Taira.
\newblock Super-resolution analysis via machine learning: A survey for fluid
  flows.
\newblock {\em arXiv preprint arXiv:2301.10937}, 2023.

\bibitem{Fathietal20}
M.~F. Fathi, I.~Perez-Raya, A.~Baghaie, P.~Berg, G.~Janiga, A.~Arzani, and
  R.~M. D'Souza.
\newblock Super-resolution and denoising of {4D-Flow MRI} using
  physics-informed deep neural nets.
\newblock {\em Computer Methods and Programs in Biomedicine}, page 105729,
  2020.

\bibitem{champion2019data}
K.~Champion, B.~Lusch, J.~N. Kutz, and S.~L. Brunton.
\newblock Data-driven discovery of coordinates and governing equations.
\newblock {\em Proceedings of the National Academy of Sciences},
  116(45):22445--22451, 2019.

\bibitem{duraisamy2021perspectives}
K.~Duraisamy.
\newblock Perspectives on machine learning-augmented reynolds-averaged and
  large eddy simulation models of turbulence.
\newblock {\em Physical Review Fluids}, 6(5):050504, 2021.

\bibitem{de2020transfer}
S.~De, J.~Britton, M.~Reynolds, R.~Skinner, K.~Jansen, and A.~Doostan.
\newblock On transfer learning of neural networks using bi-fidelity data for
  uncertainty propagation.
\newblock {\em International Journal for Uncertainty Quantification}, 10(6),
  2020.

\bibitem{shukla2023deep}
K.~Shukla, V.~Oommen, A.~Peyvan, M.~Penwarden, L.~Bravo, A.~Ghoshal, R.~M.
  Kirby, and G.~E. Karniadakis.
\newblock Deep neural operators can serve as accurate surrogates for shape
  optimization: a case study for airfoils.
\newblock {\em arXiv preprint arXiv:2302.00807}, 2023.

\bibitem{yuan2022towards}
L.~Yuan, H.~S. Park, and E.~Lejeune.
\newblock Towards out of distribution generalization for problems in mechanics.
\newblock {\em Computer Methods in Applied Mechanics and Engineering},
  400:115569, 2022.

\bibitem{kutz2022parsimony}
J.~N. Kutz and S.~L. Brunton.
\newblock Parsimony as the ultimate regularizer for physics-informed machine
  learning.
\newblock {\em Nonlinear Dynamics}, 107(3):1801--1817, 2022.

\bibitem{oh2023genetic}
H.~Oh, R.~Amici, G.~Bomarito, S.~Zhe, R.~Kirby, and J.~Hochhalter.
\newblock Genetic programming based symbolic regression for analytical
  solutions to differential equations.
\newblock {\em arXiv preprint arXiv:2302.03175}, 2023.

\bibitem{brunton2016discovering}
S.~L. Brunton, J.~L. Proctor, and J.~N. Kutz.
\newblock Discovering governing equations from data by sparse identification of
  nonlinear dynamical systems.
\newblock {\em Proceedings of the National Academy of Sciences},
  113(15):3932--3937, 2016.

\bibitem{kapteyn2020toward}
M.~G. Kapteyn, D.~J. Knezevic, and K.~Willcox.
\newblock Toward predictive digital twins via component-based reduced-order
  models and interpretable machine learning.
\newblock In {\em AIAA Scitech 2020 Forum}, page 0418, 2020.

\bibitem{samek2021explaining}
W.~Samek, G.~Montavon, S.~Lapuschkin, C.~J. Anders, and K.~R. M{\"u}ller.
\newblock Explaining deep neural networks and beyond: A review of methods and
  applications.
\newblock {\em Proceedings of the IEEE}, 109(3):247--278, 2021.

\bibitem{thampi2022interpretable}
A.~Thampi.
\newblock {\em Interpretable {AI}: Building explainable machine learning
  systems}.
\newblock Simon and Schuster, 2022.

\bibitem{zhong2022explainable}
X.~Zhong, B.~Gallagher, S.~Liu, B.~Kailkhura, A.~Hiszpanski, and T~Y.~J. Han.
\newblock Explainable machine learning in materials science.
\newblock {\em Npj Computational Materials}, 8(1):204, 2022.

\bibitem{rasheed2022explainable}
K.~Rasheed, A.~Qayyum, M.~Ghaly, A.~Al-Fuqaha, A.~Razi, and J.~Qadir.
\newblock Explainable, trustworthy, and ethical machine learning for
  healthcare: A survey.
\newblock {\em Computers in Biology and Medicine}, page 106043, 2022.

\bibitem{salahuddin2022transparency}
Z.~Salahuddin, H.~C. Woodruff, A.~Chatterjee, and P.~Lambin.
\newblock Transparency of deep neural networks for medical image analysis: A
  review of interpretability methods.
\newblock {\em Computers in Biology and Medicine}, 140:105111, 2022.

\bibitem{sutthithatip2021explainable}
S.~Sutthithatip, S.~Perinpanayagam, S.~Aslam, and A.~Wileman.
\newblock Explainable {AI} in aerospace for enhanced system performance.
\newblock In {\em 2021 {IEEE/AIAA} 40th Digital Avionics Systems Conference
  ({DASC})}, pages 1--7. IEEE, 2021.

\bibitem{saez2022convolutional}
H.~S{\'a}ez~de Oc{\'a}riz~Borde, D.~Sondak, and P.~Protopapas.
\newblock Convolutional neural network models and interpretability for the
  anisotropic {Reynolds} stress tensor in turbulent one-dimensional flows.
\newblock {\em Journal of Turbulence}, 23(1-2):1--28, 2022.

\bibitem{kim2023interpretable}
H.~Kim, J.~Kim, and C.~Lee.
\newblock Interpretable deep learning for prediction of {Prandtl} number effect
  in turbulent heat transfer.
\newblock {\em Journal of Fluid Mechanics}, 955:A14, 2023.

\bibitem{cremades2023explaining}
A.~Cremades, S.~Hoyas, P.~Quintero, M.~Lellep, M.~Linkmann, and R.~Vinuesa.
\newblock Explaining wall-bounded turbulence through deep learning.
\newblock {\em arXiv preprint arXiv:2302.01250}, 2023.

\bibitem{fang2022data}
L.~Fang, T.~W. Bao, W.~Q. Xu, Z.~D. Zhou, J.~L. Du, and Y.~Jin.
\newblock Data driven turbulence modeling in turbomachinery--an applicability
  study.
\newblock {\em Computers \& Fluids}, 238:105354, 2022.

\bibitem{muckley2022interpretable}
E.~S. Muckley, J.~E. Saal, B.~Meredig, C.~S. Roper, and J.~H. Martin.
\newblock Interpretable models for extrapolation in scientific machine
  learning.
\newblock {\em arXiv preprint arXiv:2212.10283}, 2022.

\bibitem{sudjianto2021designing}
A.~Sudjianto and A.~Zhang.
\newblock Designing inherently interpretable machine learning models.
\newblock {\em arXiv preprint arXiv:2111.01743}, 2021.

\bibitem{horvath2012inference}
L.~Horv{\'a}th and P.~Kokoszka.
\newblock {\em Inference for functional data with applications}, volume 200.
\newblock Springer Science \& Business Media, 2012.

\bibitem{wang2016functional}
J.~L. Wang, J.~M. Chiou, and H.~G. M{\"u}ller.
\newblock Functional data analysis.
\newblock {\em Annual Review of Statistics and its application}, 3:257--295,
  2016.

\bibitem{ullah2013applications}
S.~Ullah and C.~F. Finch.
\newblock Applications of functional data analysis: A systematic review.
\newblock {\em BMC Medical Research Methodology}, 13:1--12, 2013.

\bibitem{arzani2022machine}
A.~Arzani, J.~X. Wang, M.~S. Sacks, and S.~C. Shadden.
\newblock Machine learning for cardiovascular biomechanics modeling: challenges
  and beyond.
\newblock {\em Annals of Biomedical Engineering}, 50(6):615--627, 2022.

\bibitem{borggaard1992optimal}
C.~Borggaard and H.~H. Thodberg.
\newblock Optimal minimal neural interpretation of spectra.
\newblock {\em Analytical Chemistry}, 64(5):545--551, 1992.

\bibitem{griswold2008hypothesis}
C.~K. Griswold, R.~Gomulkiewicz, and N.~Heckman.
\newblock Hypothesis testing in comparative and experimental studies of
  function-valued traits.
\newblock {\em Evolution}, 62(5):1229--1242, 2008.

\bibitem{ferratyoxford}
F.~Ferraty and Y.~Romain.
\newblock The {Oxford} handbook of functional data analysis, 2011.

\bibitem{kohler2014review}
M.~K{\"o}hler, A.~Schindler, and S.~Sperlich.
\newblock A review and comparison of bandwidth selection methods for kernel
  regression.
\newblock {\em International Statistical Review}, 82(2):243--274, 2014.

\bibitem{ghosh2018kernel}
S.~Ghosh.
\newblock {\em Kernel smoothing: Principles, methods and applications}.
\newblock John Wiley \& Sons, 2018.

\bibitem{csala2022comparing}
H.~Csala, S.~Dawson, and A.~Arzani.
\newblock Comparing different nonlinear dimensionality reduction techniques for
  data-driven unsteady fluid flow modeling.
\newblock {\em Physics of Fluids}, 34(11), 2022.

\bibitem{baddoo2022kernel}
P.~J. Baddoo, B.~Herrmann, B.~J. McKeon, and S.~L. Brunton.
\newblock Kernel learning for robust dynamic mode decomposition: linear and
  nonlinear disambiguation optimization.
\newblock {\em Proceedings of the Royal Society A}, 478(2260):20210830, 2022.

\bibitem{kovachki2023neural}
N.~Kovachki, Z.~Li, B.~Liu, K.~Azizzadenesheli, K.~Bhattacharya, A.~Stuart, and
  A.~Anandkumar.
\newblock Neural operator: Learning maps between function spaces with
  applications to {PDEs}.
\newblock {\em Journal of Machine Learning Research}, 24(89):1--97, 2023.

\bibitem{li2020fourier}
Z.~Li, N.~Kovachki, K.~Azizzadenesheli, B.~Liu, K.~Bhattacharya, A.~Stuart, and
  A.~Anandkumar.
\newblock Fourier neural operator for parametric partial differential
  equations.
\newblock {\em arXiv preprint arXiv:2010.08895}, 2020.

\bibitem{yin2022simulating}
M.~Yin, E.~Ban, B.~V. Rego, E.~Zhang, C.~Cavinato, J.~D. Humphrey, and
  G.~Em~Karniadakis.
\newblock Simulating progressive intramural damage leading to aortic dissection
  using {DeepONet}: an operator--regression neural network.
\newblock {\em Journal of the Royal Society Interface}, 19(187):20210670, 2022.

\bibitem{you2022physics}
H.~You, Q.~Zhang, C.~J. Ross, C.~H. Lee, M.~C. Hsu, and Y.~Yu.
\newblock A physics-guided neural operator learning approach to model
  biological tissues from digital image correlation measurements.
\newblock {\em Journal of Biomechanical Engineering}, 144(12):121012, 2022.

\bibitem{renn2023forecasting}
P.~I. Renn, C.~Wang, S.~Lale, Z.~Li, A.~Anandkumar, and M.~Gharib.
\newblock Forecasting subcritical cylinder wakes with {Fourier Neural
  Operators}.
\newblock {\em arXiv preprint arXiv:2301.08290}, 2023.

\bibitem{li2020neural}
Z.~Li, N.~Kovachki, K.~Azizzadenesheli, B.~Liu, K.~Bhattacharya, A.~Stuart, and
  A.~Anandkumar.
\newblock Neural operator: Graph kernel network for partial differential
  equations.
\newblock {\em arXiv preprint arXiv:2003.03485}, 2020.

\bibitem{duffy2015green}
D.~G. Duffy.
\newblock {\em Green's functions with applications}.
\newblock CRC press, 2015.

\bibitem{nair2011advanced}
S.~Nair.
\newblock {\em Advanced topics in applied mathematics: for engineering and the
  physical sciences}.
\newblock Cambridge University Press, 2011.

\bibitem{voulodimos2018deep}
A.~Voulodimos, N.~Doulamis, A.~Doulamis, and E.~Protopapadakis.
\newblock Deep learning for computer vision: A brief review.
\newblock {\em Computational Intelligence and Neuroscience}, 2018, 2018.

\bibitem{pandey2020perspective}
S.~Pandey, J.~Schumacher, and K.~R. Sreenivasan.
\newblock A perspective on machine learning in turbulent flows.
\newblock {\em Journal of Turbulence}, 21(9-10):567--584, 2020.

\bibitem{guastoni2021convolutional}
L.~Guastoni, A.~G{\"u}emes, A.~Ianiro, S.~Discetti, P.~Schlatter, H.~Azizpour,
  and R.~Vinuesa.
\newblock Convolutional-network models to predict wall-bounded turbulence from
  wall quantities.
\newblock {\em Journal of Fluid Mechanics}, 928:A27, 2021.

\bibitem{zhang2021tonr}
Z.~Zhang, Y.~Li, W.~Zhou, X.~Chen, W.~Yao, and Y.~Zhao.
\newblock {TONR}: An exploration for a novel way combining neural network with
  topology optimization.
\newblock {\em Computer Methods in Applied Mechanics and Engineering},
  386:114083, 2021.

\bibitem{naylor1982linear}
A.~W. Naylor and G.~R. Sell.
\newblock {\em Linear operator theory in engineering and science}.
\newblock Springer Science \& Business Media, 1982.

\bibitem{Aggarwal18}
C.~C. Aggarwal.
\newblock {\em Neural Networks and Deep Learning: A Textbook}.
\newblock Springer, 2018.

\bibitem{williams2006gaussian}
C.~K.~I. Williams and C.~E. Rasmussen.
\newblock {\em Gaussian processes for machine learning}, volume~2.
\newblock MIT press Cambridge, MA, 2006.

\bibitem{neal2012bayesian}
R.~M. Neal.
\newblock {\em Bayesian learning for neural networks}, volume 118.
\newblock Springer Science \& Business Media, 2012.

\bibitem{goswami2022physics}
S.~Goswami, A.~Bora, Y.~Yu, and G.~E. Karniadakis.
\newblock Physics-informed neural operators.
\newblock {\em arXiv preprint arXiv:2207.05748}, 2022.

\bibitem{huang2023introduction}
O.~Huang, S.~Saha, J.~Guo, and W.~K. Liu.
\newblock An introduction to kernel and operator learning methods for
  homogenization by self-consistent clustering analysis.
\newblock {\em Computational Mechanics}, pages 1--25, 2023.

\bibitem{qian2020lift}
El. Qian, B.~Kramer, B.~Peherstorfer, and K.~Willcox.
\newblock Lift \& learn: Physics-informed machine learning for large-scale
  nonlinear dynamical systems.
\newblock {\em Physica D: Nonlinear Phenomena}, 406:132401, 2020.

\bibitem{muller2005generalized}
H.~G. M{\"u}ller and U.~Stadtm{\"u}ller.
\newblock Generalized functional linear models.
\newblock {\em Annals of Statistics}, 33(2):774--805, 2005.

\bibitem{horova2012kernel}
I.~Horov{\'a}, J.~Kolacek, and J.~Zelinka.
\newblock {\em Kernel Smoothing in {MATLAB}: theory and practice of kernel
  smoothing}.
\newblock World scientific, 2012.

\bibitem{muller2008functional}
H.~G. M{\"u}ller and F.~Yao.
\newblock Functional additive models.
\newblock {\em Journal of the American Statistical Association},
  103(484):1534--1544, 2008.

\bibitem{agarwal2021neural}
R.~Agarwal, L.~Melnick, N.~Frosst, X.~Zhang, B.~Lengerich, R.~Caruana, and
  G.~E. Hinton.
\newblock Neural additive models: Interpretable machine learning with neural
  nets.
\newblock {\em Advances in Neural Information Processing Systems},
  34:4699--4711, 2021.

\bibitem{james2009functional}
G.~M. James, J.~Wang, and J.~Zhu.
\newblock Functional linear regression that's interpretable.
\newblock {\em The Annals of Statistics}, 37(5A):2083--2108, 2009.

\bibitem{marcinkevivcs2023interpretable}
R.~Marcinkevi{\v{c}}s and J.~E. Vogt.
\newblock Interpretable and explainable machine learning: A methods-centric
  overview with concrete examples.
\newblock {\em Wiley Interdisciplinary Reviews: Data Mining and Knowledge
  Discovery}, 13(3):e1493, 2023.

\bibitem{xu2023sparse}
S.~Xu, Z.~Bu, P.~Chaudhari, and I.~J. Barnett.
\newblock Sparse neural additive model: Interpretable deep learning with
  feature selection via group sparsity.
\newblock In {\em Joint European Conference on Machine Learning and Knowledge
  Discovery in Databases}, pages 343--359. Springer, 2023.

\bibitem{molnar2020quantifying}
C.~Molnar, G.~Casalicchio, and B.~Bischl.
\newblock Quantifying model complexity via functional decomposition for better
  post-hoc interpretability.
\newblock In {\em Machine Learning and Knowledge Discovery in Databases:
  International Workshops of ECML PKDD 2019}, pages 193--204. Springer, 2020.

\bibitem{devries2018learning}
T.~DeVries and G.~W. Taylor.
\newblock Learning confidence for out-of-distribution detection in neural
  networks.
\newblock {\em arXiv preprint arXiv:1802.04865}, 2018.

\bibitem{yang2021generalized}
J.~Yang, K.~Zhou, Y.~Li, and Z.~Liu.
\newblock Generalized out-of-distribution detection: A survey.
\newblock {\em arXiv preprint arXiv:2110.11334}, 2021.

\bibitem{yuan2022mechanical}
L.~Yuan, H.~S. Park, and E.~Lejeune.
\newblock Mechanical {MNIST}--distribution shift.
\newblock 2022.

\bibitem{lecun1998gradient}
Y.~LeCun, L.~Bottou, Y.~Bengio, and P.~Haffner.
\newblock Gradient-based learning applied to document recognition.
\newblock {\em Proceedings of the IEEE}, 86(11):2278--2324, 1998.

\bibitem{cohen2017emnist}
G.~Cohen, S.~Afshar, J.~Tapson, and A.~Van~Schaik.
\newblock {EMNIST}: Extending {MNIST} to handwritten letters.
\newblock In {\em 2017 international joint conference on neural networks
  (IJCNN)}, pages 2921--2926. IEEE, 2017.

\bibitem{lejeune2020mechanical}
E.~Lejeune.
\newblock Mechanical {MNIST}: A benchmark dataset for mechanical metamodels.
\newblock {\em Extreme Mechanics Letters}, 36:100659, 2020.

\bibitem{LoggMardalWells12}
A.~Logg, K.~A. Mardal, and G.~Wells.
\newblock {\em Automated solution of differential equations by the finite
  element method}, volume~84.
\newblock Springer, Berlin, Heidelberg, 2012.

\bibitem{ArzaniWangDsouza21}
A.~Arzani, J.~X. Wang, and R.~M. D'Souza.
\newblock Uncovering near-wall blood flow from sparse data with
  physics-informed neural networks.
\newblock {\em Physics of Fluids}, 33(7), 2021.

\bibitem{Aliakbarietal23}
M.~Aliakbari, M.~S. Sadrabadi, P.~Vadasz, and A.~Arzani.
\newblock Ensemble physics informed neural networks: A framework to improve
  inverse transport modeling in heterogeneous domains.
\newblock {\em Physics of Fluids}, 33:053616, 2023.

\bibitem{de2022neural}
S.~De and A.~Doostan.
\newblock Neural network training using {L1-regularization} and bi-fidelity
  data.
\newblock {\em Journal of Computational Physics}, 458:111010, 2022.

\bibitem{aliakbari2022predicting}
M.~Aliakbari, M.~Mahmoudi, P.~Vadasz, and A.~Arzani.
\newblock Predicting high-fidelity multiphysics data from low-fidelity fluid
  flow and transport solvers using physics-informed neural networks.
\newblock {\em International Journal of Heat and Fluid Flow}, 96:109002, 2022.

\bibitem{altman1995bandwidth}
N.~Altman and C.~Leger.
\newblock Bandwidth selection for kernel distribution function estimation.
\newblock {\em Journal of Statistical Planning and Inference}, 46(2):195--214,
  1995.

\bibitem{zhu2023reliable}
M.~Zhu, H.~Zhang, A.~Jiao, G.~E. Karniadakis, and L.~Lu.
\newblock Reliable extrapolation of deep neural operators informed by physics
  or sparse observations.
\newblock {\em Computer Methods in Applied Mechanics and Engineering},
  412:116064, 2023.

\bibitem{lemhadri2021lassonet}
I.~Lemhadri, F.~Ruan, L.~Abraham, and R.~Tibshirani.
\newblock Lassonet: A neural network with feature sparsity.
\newblock {\em The Journal of Machine Learning Research}, 22(1):5633--5661,
  2021.

\bibitem{gin2021deepgreen}
C.~R. Gin, D.~E. Shea, S.~L. Brunton, and J.~N. Kutz.
\newblock Deepgreen: deep learning of {Green's} functions for nonlinear
  boundary value problems.
\newblock {\em Scientific Reports}, 11(1):21614, 2021.

\bibitem{boulle2022data}
N.~Boull{\'e}, C.~J. Earls, and A.~Townsend.
\newblock Data-driven discovery of {Green's} functions with
  human-understandable deep learning.
\newblock {\em Scientific Reports}, 12(1):4824, 2022.

\bibitem{budivsic2012applied}
M.~Budi{\v{s}}i{\'c}, R.~Mohr, and I.~Mezi{\'c}.
\newblock Applied koopmanism.
\newblock {\em Chaos: An Interdisciplinary Journal of Nonlinear Science},
  22(4):047510, 2012.

\bibitem{mezic2013analysis}
I.~Mezi{\'c}.
\newblock Analysis of fluid flows via spectral properties of the {Koopman}
  operator.
\newblock {\em Annual Review of Fluid Mechanics}, 45:357--378, 2013.

\bibitem{takeishi2017learning}
N.~Takeishi, Y.~Kawahara, and T.~Yairi.
\newblock Learning koopman invariant subspaces for dynamic mode decomposition.
\newblock {\em Advances in neural information processing systems}, 30, 2017.

\bibitem{jovanovic2014sparsity}
M.~R. Jovanovi{\'c}, P.~J. Schmid, and J.~W. Nichols.
\newblock Sparsity-promoting dynamic mode decomposition.
\newblock {\em Physics of Fluids}, 26(2):024103, 2014.

\bibitem{flaschel2021unsupervised}
M.~Flaschel, S.~Kumar, and L.~De~Lorenzis.
\newblock Unsupervised discovery of interpretable hyperelastic constitutive
  laws.
\newblock {\em Computer Methods in Applied Mechanics and Engineering},
  381:113852, 2021.

\bibitem{BrightLinKutz13}
I.~Bright, G.~Lin, and J.~N. Kutz.
\newblock Compressive sensing based machine learning strategy for
  characterizing the flow around a cylinder with limited pressure measurements.
\newblock {\em Physics of Fluids}, 25(12):127102, 2013.

\bibitem{ArzaniDawson21}
A.~Arzani and S.~Dawson.
\newblock Data-driven cardiovascular flow modelling: examples and
  opportunities.
\newblock {\em Journal of The Royal Society Interface}, 18:20200802, 2021.

\bibitem{li2010generalized}
Y.~Li, N.~Wang, and R.~J. Carroll.
\newblock Generalized functional linear models with semiparametric single-index
  interactions.
\newblock {\em Journal of the American Statistical Association},
  105(490):621--633, 2010.

\bibitem{panton2013incompressible}
R.~L. Panton.
\newblock {\em Incompressible flow}.
\newblock John Wiley \& Sons, 2013.

\bibitem{xiong2023neural}
S.~Xiong, X.~He, Y.~Tong, Y.~Deng, and B.~Zhu.
\newblock Neural vortex method: from finite {Lagrangian} particles to infinite
  dimensional {Eulerian} dynamics.
\newblock {\em Computers \& Fluids}, 258:105811, 2023.

\bibitem{isakov2006inverse}
V.~Isakov.
\newblock {\em Inverse problems for partial differential equations}, volume
  127.
\newblock Springer, 2006.

\bibitem{delillo2003detection}
T.~DeLillo, V.~Isakov, N.~Valdivia, and L.~Wang.
\newblock The detection of surface vibrations from interior acoustical
  pressure.
\newblock {\em Inverse Problems}, 19(3):507, 2003.

\bibitem{lu2021learning}
L.~Lu, P.~Jin, G.~Pang, Z.~Zhang, and George~E. Karniadakis.
\newblock Learning nonlinear operators via {DeepONet} based on the universal
  approximation theorem of operators.
\newblock {\em Nature Machine Intelligence}, 3(3):218--229, 2021.

\bibitem{boulle2022learning}
N.~Boull{\'e}, S.~Kim, T.~Shi, and A.~Townsend.
\newblock Learning green's functions associated with time-dependent partial
  differential equations.
\newblock {\em Journal of Machine Learning Research}, 23(218):1--34, 2022.

\bibitem{liang2023synergistic}
L.~Liang, M.~Liu, J.~Elefteriades, and W.~Sun.
\newblock Synergistic integration of deep neural networks and finite element
  method with applications for biomechanical analysis of human aorta.
\newblock {\em bioRxiv}, pages 2023--04, 2023.

\bibitem{woolley2003p}
T.~W. Woolley.
\newblock The {p-value, the Bayes/Neyman-Pearson Compromise} and the teaching
  of statistical inference in introductory business statistics.
\newblock {\em Proceedings of the Academy of Business Education}, 4:823, 2003.

\end{thebibliography}

\end{document}